\documentclass[10pt,twocolumn,letterpaper]{article}

\usepackage{subcaption}
\usepackage{cvpr}
\usepackage{times}
\usepackage{epsfig}
\usepackage{graphicx}
\usepackage{amsmath}
\usepackage{amssymb}
\usepackage{algorithm}
\usepackage{algpseudocode}
\usepackage{booktabs}
\usepackage{multirow}

\usepackage[breaklinks=true,bookmarks=false]{hyperref}

\cvprfinalcopy 

\begin{document}

\title{Pay attention to the activations: a modular attention mechanism \\for fine-grained image recognition}

\author{Pau Rodr\'iguez\\
Element AI \\
Montreal, Canada \\
{\tt\small pau.rodriguez@elementai.com}
\and
Diego Velazquez\\
Computer Vision Center\\
Barcelona, Spain\\
{}
\and
Guillem Cucurull \\
Element AI\\
Montreal, Canada\\
{}
\and
Josep M. Gonfaus \\
Visual Tagging Services\\
Barcelona, Spain\\
{}
\and
F. Xavier Roca\\
Universitat Autonoma de Bacelona\\
Barcelona, Spain\\
{}
\and
Jordi Gonz\`alez \\
Universitat Autonoma de Barcelona\\
Barcelona, Spain\\
{}
}

\maketitle

\begin{abstract}
Fine-grained image recognition is central to many multimedia tasks such as search, retrieval and captioning. Unfortunately, these tasks are still challenging since the appearance of samples of the same class can be more different than those from different classes. This issue is mainly due to changes in deformation, pose, and the presence of clutter. In the literature, attention has been one of the most successful strategies to handle the aforementioned problems. Attention has been typically implemented in neural networks by selecting the most informative regions of the image that improve classification. In contrast, in this paper, attention is not applied at the image level but to the convolutional feature activations. In essence, with our approach, the neural model learns to attend to lower-level feature activations without requiring part annotations and uses those activations to update and rectify the output likelihood distribution. The proposed mechanism is modular, architecture-independent and efficient in terms of both parameters and computation required. Experiments demonstrate that well-known networks such as Wide Residual Networks and ResNeXt, when augmented with our approach, systematically improve their classification accuracy and become more robust to changes in deformation and pose and to the presence of clutter. As a result, our proposal reaches state-of-the-art classification accuracies in CIFAR-10, the Adience gender recognition task, Stanford Dogs, and UEC-Food100 while obtaining competitive performance in ImageNet, CIFAR-100, CUB200 Birds, and Stanford Cars. In addition, we analyze the different components of our model, showing that the proposed attention modules succeed in finding the most discriminative regions of the image. Finally, as a proof of concept, we demonstrate that with only local predictions, an augmented neural network can successfully classify an image before reaching any fully connected layer, thus reducing the computational amount up to 10\%.   
\end{abstract}

\begin{figure*}[!t!]
\centering 
	\includegraphics[width=\textwidth]{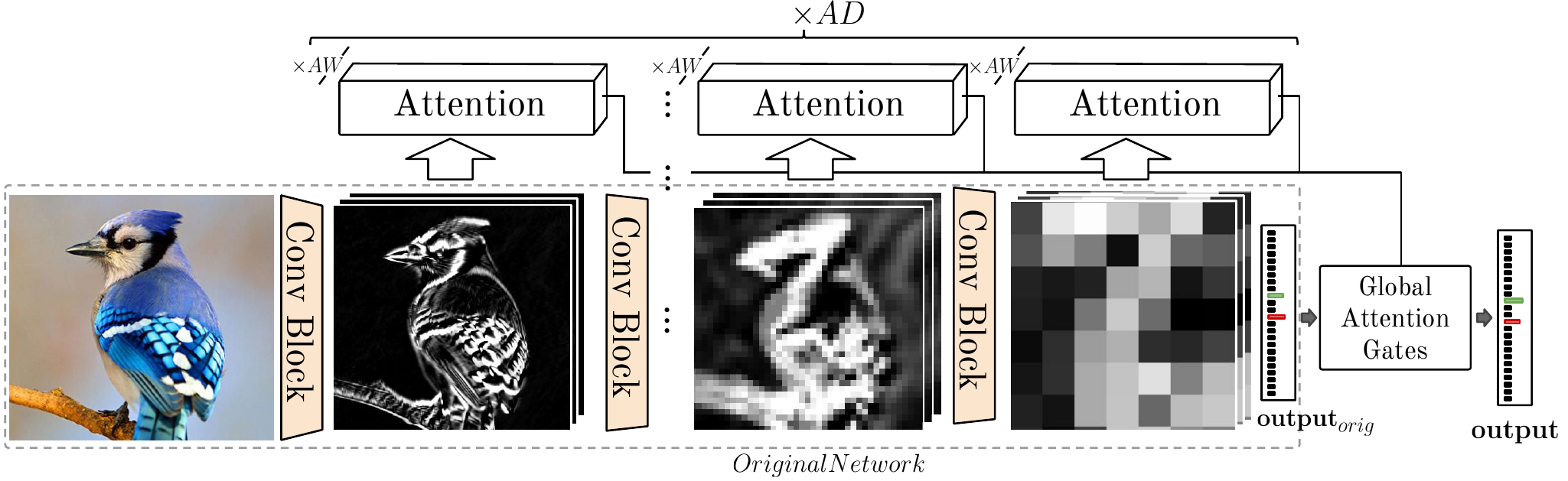}
\caption{The proposed approach. Each of the $AD$ (attention depth) attention modules uses $AW$ (attention width) attention heads to produce local class predictions at different network depths. These predictions correct the original network output ($\mathbf{output}_{orig}$) with the global attention gates }
\label{fig:overall}
\end{figure*}

{\let\thefootnote\relax\footnote{{\textcopyright~2019 IEEE.  Personal use of this material is permitted.  Permission from IEEE must be obtained for all other uses, in any current or future media, including reprinting/republishing this material for advertising or promotional purposes, creating new collective works, for resale or redistribution to servers or lists, or reuse of any copyrighted component of this work in other works.}}}
\section{Introduction}
The latest advances in computer vision and machine learning such as large-scale image recognition with convolutional neural networks (CNNs) \cite{krizhevsky2012imagenet} have greatly impacted multimedia systems, especially for content retrieval and analysis. For instance, CNNs have been used in a wide range of interesting applications such as cross-modal retrieval \cite{wang2015learning, he2016cross}, multimedia content description \cite{cho2015describing}, food recipe retrieval \cite{min2017being}, and clothing retrieval \cite{liang2016clothes}. 

However, fine-grained image recognition tasks, such as the last two previously listed, remain challenging for the current models. In these kinds of tasks, multimedia systems need to discriminate between subclasses of the same category, such as different kinds of hats or different kinds of soup. In this regime, the difference between two classes resides in subtle details that can be overwhelmed by the differences in pose, clutter, and deformation present in natural images. As a result, compared to other methods, traditional CNNs yield lower performance rates on fine-grained recognition benchmarks \cite{xie2017aggregated}.

To address this problem, researchers have drawn inspiration from human vision, which selectively focuses on the most informative regions of the image \cite{itti1998model, ungerleider2000mechanisms}. For instance, attention mechanisms have been used for captioning \cite{gao2017video, zhang2018high}, action recognition \cite{li2018unified}, person reidentification \cite{wu2018and}, and fine-grained image classification \cite{zhao2017diversified}. In \cite{rodriguez2018attend}, we proposed an efficient model to improve the image classification performance of any architecture by attending to its own feature maps. In this work, we extend \cite{rodriguez2018attend} with experiments on the Imagenet \cite{russakovsky2012imagenet}, additional architectures such as ResNeXt \cite{xie2017aggregated}, and additional insight and visualization of the attention activations, and we demonstrate that the proposed algorithm could be used for adaptive computation.

Another approach to modeling attention mechanisms is through recurrent architectures \cite{hochreiter1997long}. For instance, the authors of \cite{sermanet2014attention, mnih2014recurrent, zhao2017diversified} propose training recurrent models to iteratively integrate information from multiple \emph{glimpses} on the input image. This way, models learn to focus on the most important regions in detail while ignoring distractors.

However, the multiple iterations of recurrent models and multiple-pass architectures add a substantial computational overhead when compared to vanilla CNNs \cite{liu2016fully}, especially during training with reinforcement learning, and thus are difficult to use in large-scale settings. On the other hand, feedforward architectures such as spatial transformer networks (STNs) \cite{jaderberg2015spatial} or residual attention networks \cite{wang2017residual} perform a single bottom-up process that discards irrelevant information at each step, and they can be trained with stochastic gradient descent. This process, however, introduces large errors when fine-grained information is discarded irreversibly since it is confused with noise at the early stages of the network. In addition, most attention models condition the design of the neural architecture, and thus, they are not suitable for transfer learning of pretrained networks using fine tuning \cite{wang2017residual, mnih2014recurrent}.
Other approaches, such as \cite{jetley2018learn}, avoid the previous problem by leaving the main architecture unaltered and producing the final prediction from attended regions of the feature activations. However, this model introduces extra computing steps, resulting in additional overhead. Differently, our approach can be executed in parallel to the augmented architecture, thus making its computational cost negligible. Moreover, in Section \ref{sect:approach}, we show that even with no parallelization, our model is more efficient than other well-known architectures.

Concretely, in this work, we propose a modular feedforward attention mechanism that is fast, does not modify the main CNN architecture and is trained with SGD. Thus, the proposed model can augment any pretrained architecture such as residual neural networks (ResNets) \cite{he2016deep} or VGG \cite{simonyan2014very}, and it is designed to be executed in parallel to the main architecture, without computational time overhead. As seen in Figure \ref{fig:overall}, the proposed mechanism consists of attention modules and attention gates. Attention modules are placed at different depths of the CNN, producing class predictions based on local information, such as fine-grained details, from the main CNN feature map activations at different levels of abstraction. Then, the attention gates use these local predictions to correct the original network output class distribution. For instance, the proposed model corrects the prediction "sedan" to "police-car" after paying attention to the blue sirens.

Experiments demonstrate that competitive baseline architectures such as wide residual networks (WRNs) augmented with our attention mechanism (WARN) consistently outperform the baseline accuracies on CIFAR \cite{krizhevsky2009learning} and six different fine-grained classification benchmarks, obtaining state-of-the-art performance on CIFAR-10, Stanford Dogs \cite{khosla2011novel}, UEC Food-100 \cite{matsuda12}, and the Adience gender classification benchmark \cite{eidinger2014age}, even with pretraining. In addition, we show that the proposed attention mechanism scales to larger datasets, such as Imagenet \cite{russakovsky2012imagenet}, and to other architectures, such as ResNeXt \cite{xie2017aggregated}, systematically enhancing the base model. Finally, we show that our approach can be used for adaptive computation since it provides class predictions from the early layers of the architecture.

This is an extension of our earlier work \cite{rodriguez2018attend} in the following ways: i) We step beyond the fine-grained recognition task by addressing large-scale image recognition, improving the accuracy on the ImageNet dataset with our proposed method. ii) We demonstrate the universality of the proposed model, improving additional architectures such as ResNeXt. iii) We provide more insight about the proposed attention modules, confirming that the attention gates are designed to choose the most informative image regions so as to maximize the performance of the classifier. iv) We show that it is possible to use the early predictions from the attention modules to extend the proposed mechanism for adaptive computation. As a result, we show that it is possible to save 10\% of the computational time with only 0.3\% accuracy loss by using the attention module confidence scores as the continuation criterion. v) In new qualitative results, we show that attention heads successfully choose the most informative parts of the image, discarding the noise, thereby confirming our hypotheses. To do so, we finally present a simple yet effective attention region visualization framework that can be used to explore any attention and saliency framework proposed in the literature.

\section{Related Work}
\subsection{Attention in Multimedia Systems}
Multimedia systems require finding specific chunks of information in large quantities of data, and hence, it is crucial to make them precise and efficient. As a result, attention mechanisms have raised considerable interest in the research community in multimedia since they allow focusing on the relevant information while ignoring the clutter. In \cite{le2013visual}, the authors highlight the importance of attention mechanisms for multimedia technologies. For instance, attention is useful for finding the most important regions of a video for efficient compression \cite{doulamis1998low, li2011visual} and error correction \cite{dhondt2006flexible, boulos2009region}, image and video retargeting \cite{avidan2007seam}, image quality assessment \cite{liu2011visual}, medical imaging \cite{cavaro2010eye, venjakob2012radiologists}, and stereoscopic 3D \cite{chamaret2010adaptive}. Attention mechanisms can also be found in more recent applications such as image and video captioning \cite{gao2017video, zhang2018high}, action recognition \cite{li2018unified}, and image and video description \cite{cho2015describing, li2018gla}.

\subsection{Recurrent Attention in Computer Vision}
In an early attempt to mimic biological vision, computer vision researchers produced a number of attention mechanisms that share a similar recurrent procedure, \emph{i.e.}, use the context to predict the next attention location \cite{paletta2005q, ranzato2014learning, butko2008pomdp, denil2012learning}. However, these models did not learn the whole procedure end-to-end, and some parts are defined beforehand. In an attempt to fully teach attention models, the authors of \cite{mnih2014recurrent} proposed a recurrent model that integrates information of multiple glimpses over the image.

In fact, recurrent attention methods naturally emerge as a way to solve fine-grained recognition problems by processing the most informative regions of the image in high resolution while skipping the less informative ones. For instance, Sermanet \emph{et al.} extended the model proposed in \cite{mnih2014recurrent} for fine-grained image recognition \cite{sermanet2014attention}. All these approaches, however, are trained with reinforcement learning (RL) and thus are slower to train since they require sampling (hard attention). In an attempt to alleviate this problem, Liu \emph{et al.} proposed a greedy reward strategy to accelerate training \cite{fu2017look} as well as producing multiple part proposals in parallel that are then classified.  Similarly, in \cite{fu2017look}, the proposed model (RA-CNN) uses a ranking loss to ensure a performance increment at every iteration.

Different from RL-based approaches, recurrent deterministic models are end-to-end differentiable (soft attention), allowing the whole architecture with SGD to be trained. For instance, in \cite{sharma2015action}, LSTMs are used to weight the importance of different regions of video frames for action recognition. In the same vein, Chen \emph{et al.} proposed attending different channels and regions of a CNN to produce better image captions with an LSTM \cite{chen2017sca}. Interestingly, in \emph{diversified visual attention} (DVAN) \cite{zhao2017diversified}, Zhao \emph{et al.} extended soft attention mechanisms enforcing nonoverlapping regions on the attention proposals. Concretely, they generate multiple patches at different resolutions from the original image, which are then processed with a CNN and integrated with an LSTM that enforces nonoverlapping attention masks on them. A similar approach was introduced in \cite{zheng2017learning}, where an ensemble of classifiers was trained over multiple regions of the output features of a CNN. In \cite{yang2018learning}, the authors used a three-agent system (NTS-Net) that interacts to maximize the efficacy on the attended regions for fine-grained classification. Wu \emph{et al.} proposed using spatial LSTMs to pool bilinear CNN features \cite{lin2015bilinear} over locally attended regions \cite{wu2018deep}.

\subsection{Feedforward Attention Models}
Instead of using iterative procedures, another set of approaches proposes feedforward attention models for fine-grained recognition. For instance, Xiao \emph{et al.} used an R-CNN \cite{zhang2014part, uijlings2013selective} to produce different part proposals, which were then pruned by relevance and classified \cite{xiao2015application}. Similarly, Peng \emph{et al.} \cite{peng2018object} proposed a multistage pipeline to first extract object proposals, obtain the most salient parts, align them with clustering, and use them to improve the classification accuracy. A simpler approach was proposed in \cite{rodriguez2017age}, where a CNN processes a low-resolution version of the image to predict where to attend on a higher resolution version. Likewise, in \cite{recasens2018learning}, the authors proposed performing a high-resolution zoom on the most salient regions of the image. Jetley \emph{et al.} proposed using the output of a CNN to produce attention proposals on the previous feature maps, using them for the final classification \cite{jetley2018learn}. Although they are not iterative, these methods introduce computational bottlenecks such as a previous feedforward step to obtain a global descriptor from which to compute the attention predictions. In \cite{sun2018multi}, the authors propose to attend to multiple regions of the network output feature map and use metric learning to enforce meaningful attended features. Since this approach is computed as a postprocessing of the output, it is orthogonal to ours, and it could be placed on top of a network augmented with our approach.  However, different from our model, it cannot be computed in parallel to the original architecture.

Other approaches, such as STNs or \emph{residual attention networks} (RANs) \cite{zilly2017recurrent}, directly incorporate mechanisms to enhance the most relevant regions and discard irrelevant information inside the network architecture and thus work in a single iteration. However, errors in the early layers of the network may discard important fine-grained information, causing a large impact on the output predictions.

In this work, different from previous approaches, we present a novel attention model that is feedforward and runs in a single iteration, does not introduce additional computational bottlenecks (since attention modules do not depend on the output of the model and thus can be run in parallel), and enhances any existing architecture. The original computational path of the main architecture is preserved so that the proposed model cannot introduce new errors as in \cite{jaderberg2015spatial, zilly2017recurrent}. As a result, wide residual networks (WRNs) augmented with our approach (denoted here as WARN) attain an error rate of 3.44\% on CIFAR10, outperforming \cite{jetley2018learn} while being seven times faster without parallelization. 

\section{Proposed approach}
\label{sect:approach}
\begin{algorithm}[!t]
\caption{Attend and Rectify}\label{alg:main}
\begin{algorithmic}[1]
\Require $L$ the network depth
\Require $N$ the number of classes
\Require $AD, AW$ the attention depth and width
\Require $\mathbf{Z^l}$ the output of the $l^{th}$ layer
\Require $\mathbf{output\_orig}$ the original network outputs
\Require $\mathbf{depths} \in \mathbb{N}^{AD}$ a list with the depths where attention is placed
\Function{AttendAndRectify}{$\mathbf{Z}$, output\_net, depths}
    \State $outs \gets zeros(AD + 1, N)$ \Comment{zeros(dim1, dim2, ...) returns a zero matrix of the corresponding dims.}
    \State $global\_gates \gets zeros(1, AD + 1)$
    \For {$i$ from $0$ to $AD$ } \Comment{Iter attention modules}
    \State $l \gets depths[i]$ 
    \State $self\_att \gets zeros(1, AW)$
    \State $head\_outputs \gets zeros(AW, N)$
    \For {$k$ from 0 to $AW$} \Comment{Iter attention heads}
        \State $mask\gets AttentionHead(\mathbf{Z}^l)$
        \State $head\_outputs[k] \gets OutputHead(\mathbf{Z}^l, mask)$
        \State $self\_att[0, k] \gets SelfAttention(\mathbf{Z}^l, mask)$
    \EndFor
    \State $self\_att \gets softmax(self\_att, dim=1)$
    \State $outs[i] \gets matmul(gates, head\_outputs)$ 
\EndFor
\State $outs[AD] \gets Classifier(\mathbf{Z}^L)$ \Comment{Original output}
\State $global\_gates \gets GlobalGates(\mathbf{Z}^L)$
\State $softmax(outs, dim=1)$
\State \Return $matmul(global\_gates, outs)[0]$ 
\EndFunction
\end{algorithmic}
\end{algorithm}

\begin{figure}[t!]
\centering 
\begin{subfigure}{0.45\textwidth}
	\includegraphics[width=\textwidth]{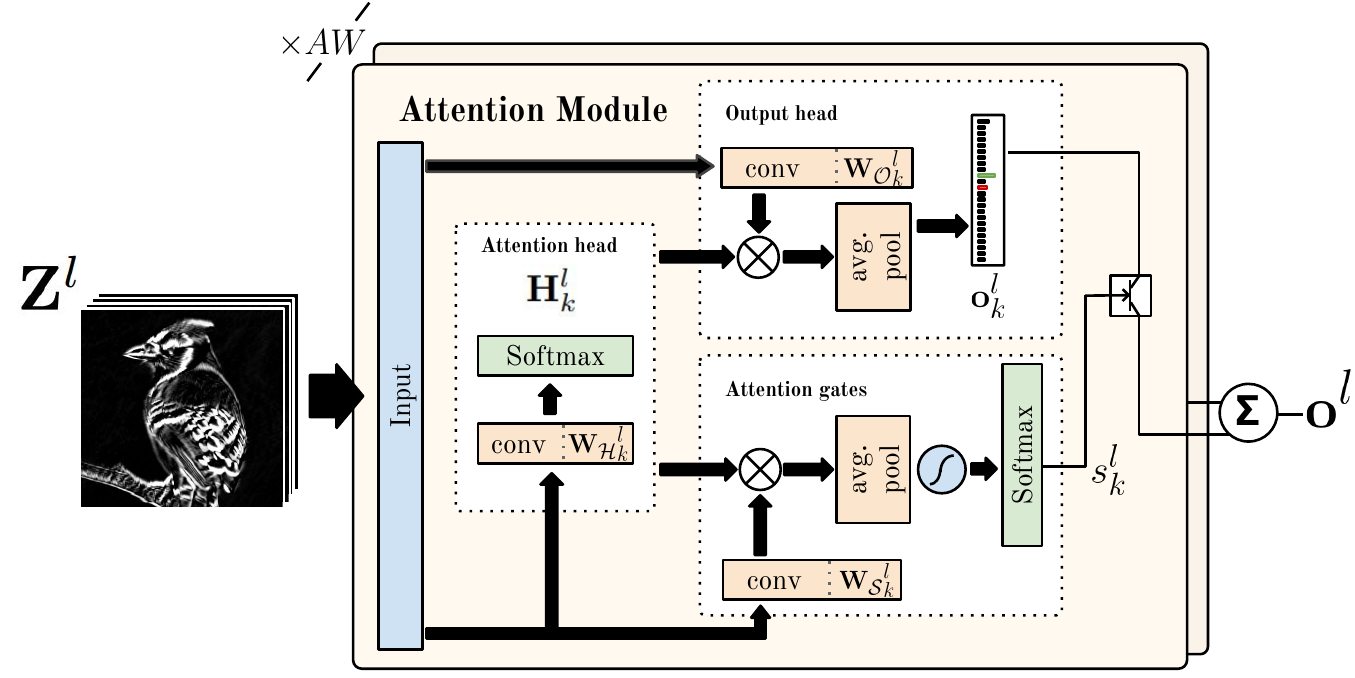}
\caption{Attention module}
\label{fig:attention_module}
\end{subfigure}
\begin{subfigure}{0.4\textwidth}
	\includegraphics[width=\textwidth]{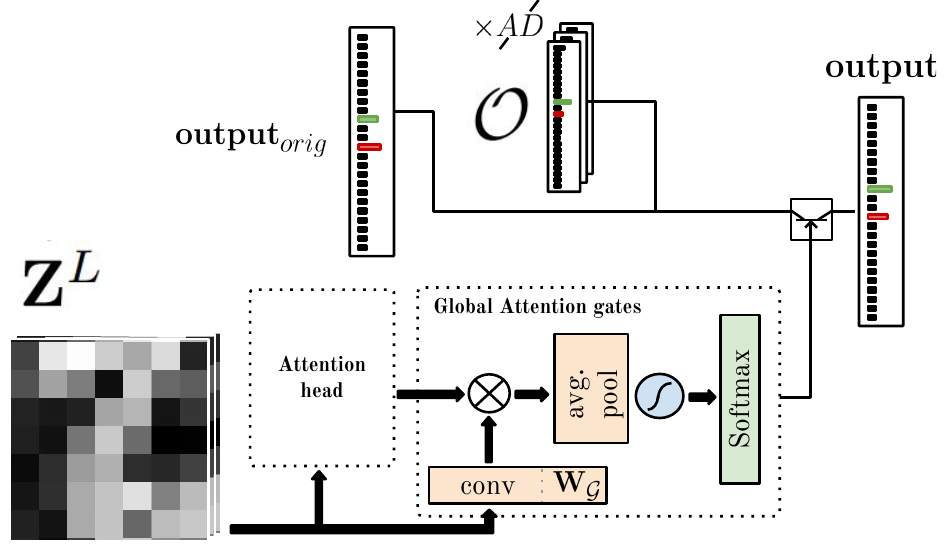}
\caption{Attention gates}
\label{fig:attention_gates}
\end{subfigure}
\caption{(a) Attention module. The feature maps $\mathbf{Z}^l$ are processed by $AW$ attention heads $\mathbf{H}^l_k$, extracting local information that is aggregated with the self-attention gates $\mathcal{S}$. (b) Global attention gates $\mathcal{G}$. These gates use global information from the last layer features $\mathbf{Z}^L$ to correct the original network $\mathbf{output}_{orig}$ with local information from the attention modules}
\label{fig:modules}
\end{figure}
As shown in Figure \ref{fig:overall}, we propose using a number of attention modules to extract local predictions at different levels of abstraction in the neural network architecture, using these to correct the original prediction of the model weighted by their respective attention gates. 
For instance, given an image of two similar bird species, the gates would choose the prediction of the attention modules that look at the most distinctive part of those species. This modular approach augments any architecture such as VGG or ResNet, without altering existing connections and with no additional supervision, \emph{i.e.}, no part annotations are used \cite{WahCUB_200_2011}. Therefore, the proposed attention modules can be used for transfer learning as they can be seamlessly included into any pretrained network.

To let the attention modules focus on multiple regions of the image, we compose for them of a set of attention heads $\mathcal{H}$. Each attention head extracts a local class prediction from a convolutional feature map, and the output of the attention module consists of the weighted average of these local predictions using self-attention gates.

The proposed mechanism is composed of the three submodules shown in Figure \ref{fig:modules}, which we describe in the next subsections: (i) attention heads $\mathcal{H}$, which learn to identify important regions in the convolutional feature activation, (ii) output heads $\mathcal{O}$, which produce a prediction using the attended regions, and (iii) self-attention gates $\mathcal{S}$ and global-attention gates $\mathcal{G}$, which weight the predictions of the multiple attention heads and modules, respectively. 

We extend the established notation of network depth and width introduced in WRNs \cite{ranzato2014learning, xie2017aggregated} by describing the attention architectures in two main dimensions: (i) \emph{attention depth} (AD), which corresponds to the number of layers augmented with our approach, and (ii) \emph{attention width} (AW), corresponding to the number of attention heads per module. This is implicitly done in the \emph{transformer} architecture \cite{vaswani2017attention} and DVAN \cite{zhao2017diversified} and provides a systematic methodology to scale attention architectures instead of defining the individual hyperparameters of each module. This way, each attention module contains the same number ($AW$) of attention heads $\mathcal{H}_k$, with $k \in [1..AW]$\footnote{Notation: $\mathcal{H,O,S,G}$ are the set of attention heads, output heads, self-attention gates, and global attention gates, respectively. Uppercase letters refer to functions or constants, and lowercase letters refer to indices. Bold uppercase letters represent matrices, and bold lowercase letters represent vectors.}; see Figure \ref{fig:overall}.

\subsection{Attention head}
 The attention heads are responsible for finding the most important regions of the image. For instance, as we show later in \ref{sec:experiments}, given a picture of a bird, the attention heads predict high scores for discriminative parts such as the beak. Concretely, each attention head receives a convolutional feature map as input and outputs a spatial heatmap with the importance of each input region:

\begin{equation}
\label{eq:atthead1}
\mathcal{H} : \mathbb{R}^{b\times c \times h \times w} \rightarrow \mathbb{R}^{b \times 1 \times h \times w},
\end{equation}

where $b$ is the batch dimension, $c$ is the number of input channels, $h$ is the feature map height and $w$ is the feature map width. To reduce the impact on the model complexity, this operation is modeled with a $1 \times 1$ convolution layer with weights $\mathbf{W}_{\mathcal{H}^l}$, where $l$ is the depth of the input feature map ($\mathbf{Z}^l$). Softmax is applied to normalize the output on the spatial dimensions so that the model learns to focus on the most important part of the image. Sigmoid units could also be employed risking activation saturation: 

\begin{equation}
\mathbf{H}^l = spatial\_softmax (\mathbf{W_\mathcal{H}}^l \ast \mathbf{Z}^l),
\label{eq:atthead2}
\end{equation}

with $\mathbf{H}^l \in \mathbb{R}^{AW\times h\times w}$ being the output matrix of the $l^{th}$ attention module, $\mathbf{W_\mathcal{H}}^l \in \mathbb{R}^{AW \times c \times h \times w}$ is a convolution kernel that produces the attention masks corresponding to the attention heads $\mathbf{H}^l$, and $\ast$ denotes the convolution operator. The softmax operation is defined as:

\begin{equation}
    softmax(\mathbf{x})_i = \frac{e^{x_i}}{\sum_{i=1}^{i=N} e^{x_i}}, \mathbf{x} \in \mathbb{R}^N
\end{equation}

We regularize $\mathbf{H}^l$ with \cite{zhao2017diversified} to prevent the attention heads of the same module from attending to the same activation region.

Note that the attention heads do not require the network output to produce their individual predictions, thus avoiding the computational bottleneck present in \cite{jetley2018learn}.

\subsection{Output head}
The output heads make local predictions based on the information selected by the attention heads. Therefore, the input is a convolutional feature map $\mathbf{Z}^l$ and the attention heads' output, and it outputs a vector with the class probabilities:

\begin{equation}
    \mathcal{O}: \mathbb{R}^{b\times c \times h \times w} \rightarrow \mathbb{R}^{b \times \#classes \times h \times w}.
\end{equation}

Similar to $\mathcal{H}$, the output heads are modeled with a $1 \times 1$ convolution kernel:

\begin{equation}
    \mathbf{o}^l_k = \frac{1}{w * h} \sum_{i=1}^{h} \sum_{j=1}^{w} [(\mathbf{W}_{\mathcal{O}^l_k} \ast \mathbf{Z}^l) \odot \mathbf{H}^l_k]_{i,j}
\end{equation},

where $
\mathbf{W_\mathcal{O}}^l_k \in \mathbb{R}^{\#classes \times c \times 1\times 1}
$ are the output head convolution weights and $\odot$ is the element-wise product.

Therefore, each $\mathbf{o}^l_{k,i,j}$ is a vector with the class probability scores predicted with the attention head $k$ at the spatial position $i \in \{1..h\}, j \in \{1..w\}$, and depth $l$.

\begin{figure*}[!t]
\centering
\begin{subfigure}[!t]{0.26\textwidth}
\centering
\includegraphics[width=0.9\textwidth]{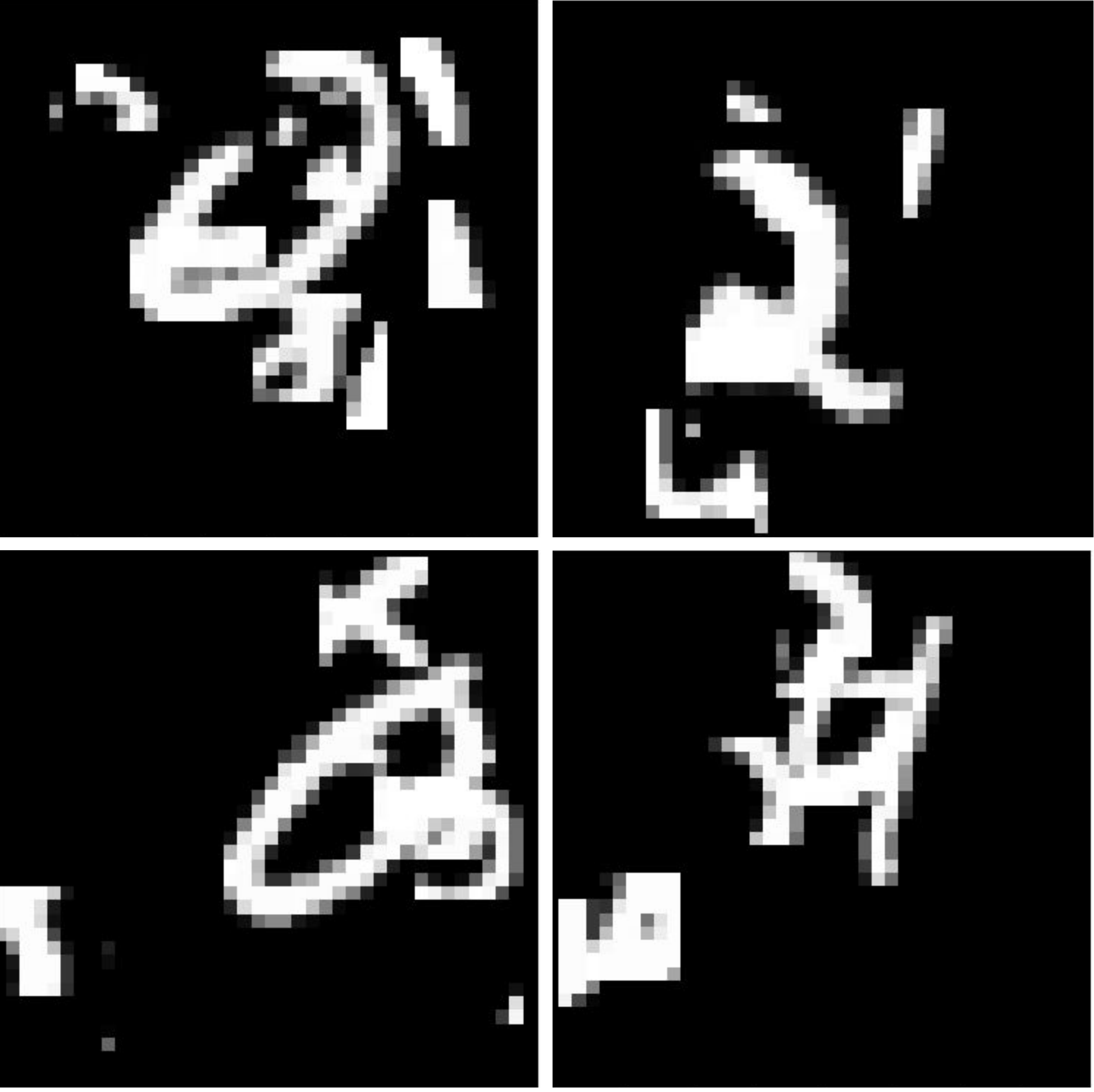}
\caption{Cluttered MNIST}
\label{fig:clutter_mnist}
\end{subfigure}
\begin{subfigure}[!t]{0.32\textwidth}
\centering
\includegraphics[width=\textwidth]{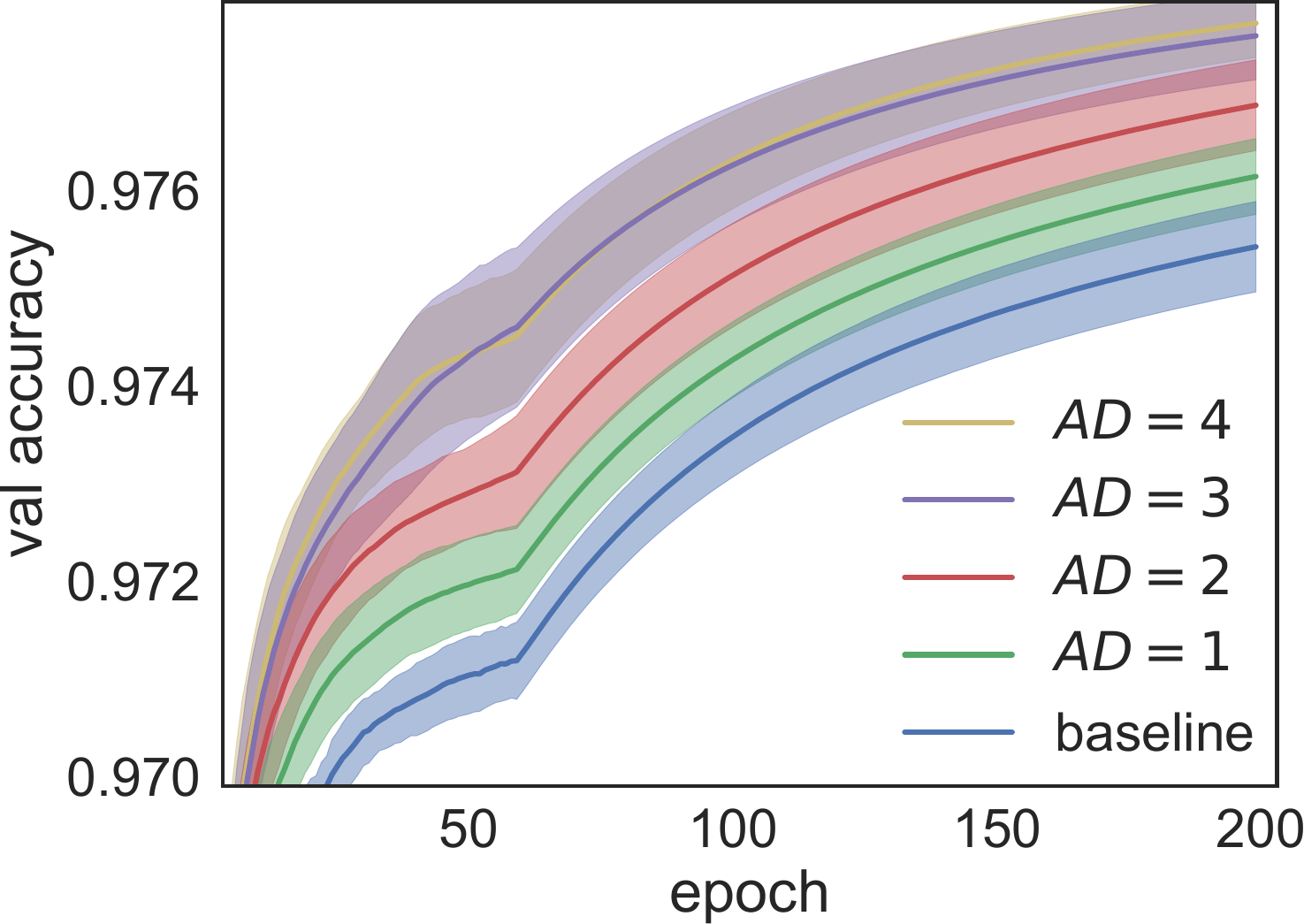}
\caption{Different depths}
\label{fig:ablation_depth}
\end{subfigure}
\begin{subfigure}[!t]{0.32\textwidth}
\centering
\includegraphics[width=\textwidth]{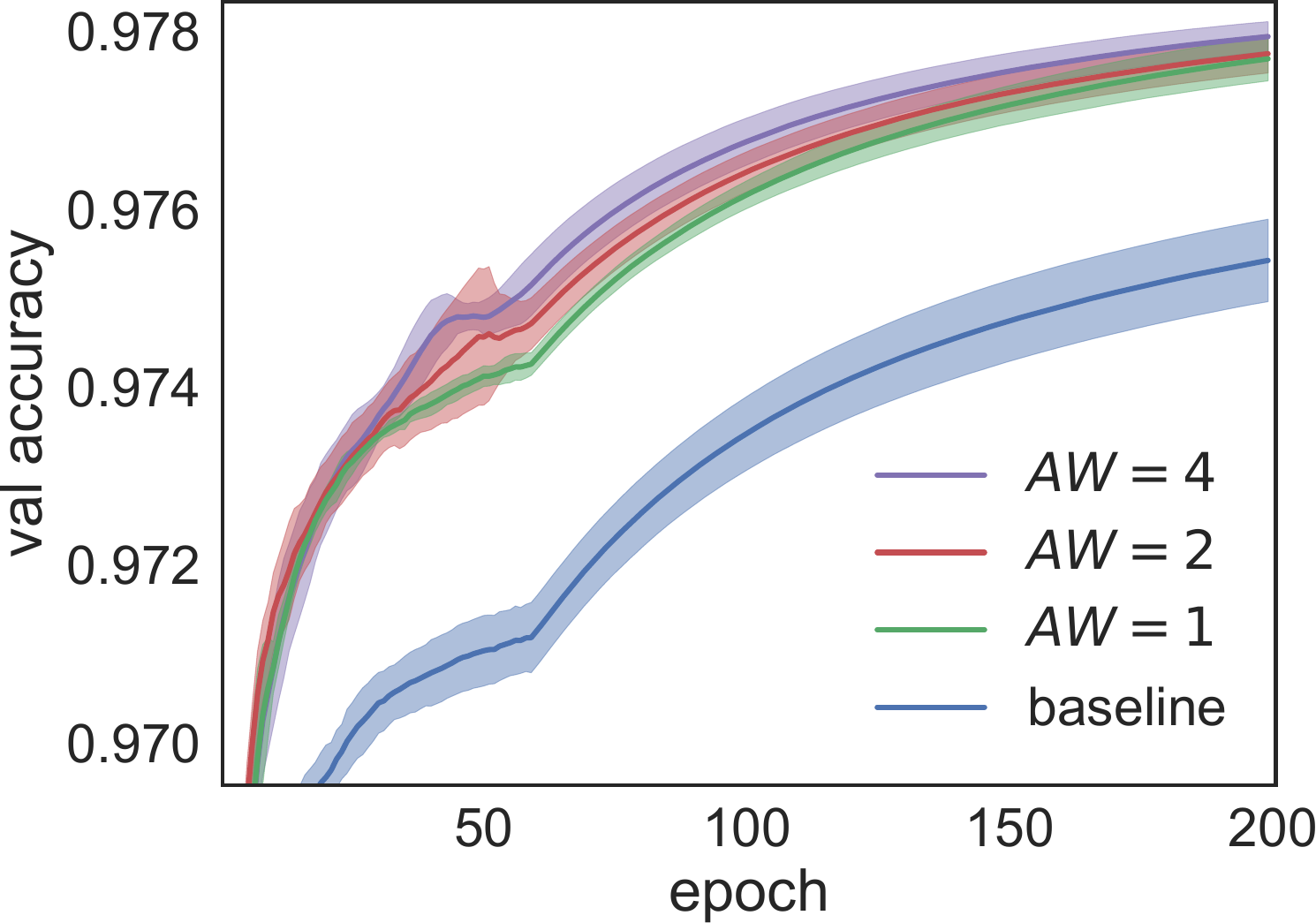}
\caption{Different widths}
\label{fig:ablation_width}
\end{subfigure}
\begin{subfigure}[!t]{0.32\textwidth}
\centering
\includegraphics[width=\textwidth]{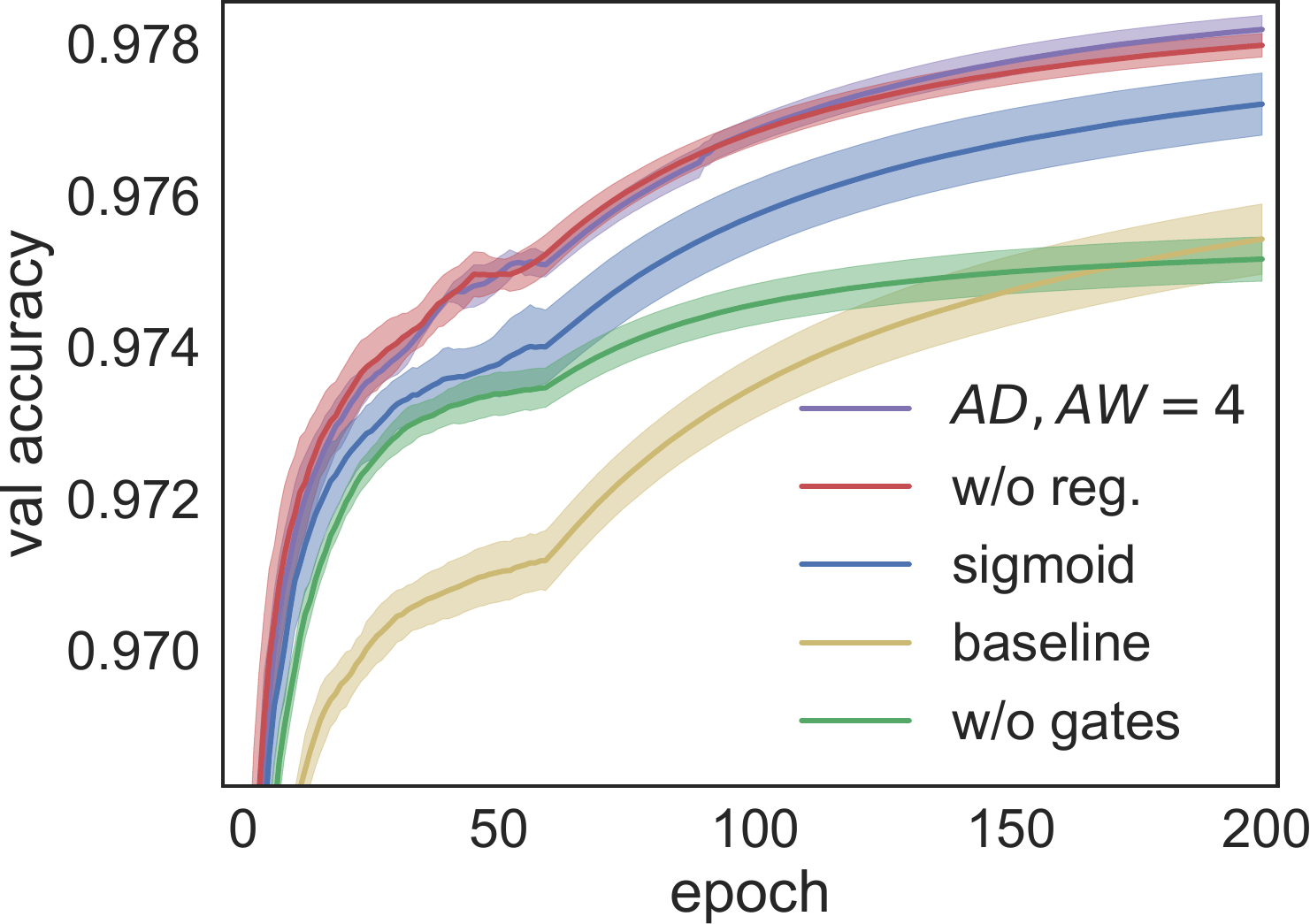}
\caption{Softmax, gates, reg.}
\label{fig:hyperparams}
\end{subfigure}
\begin{subfigure}[!t]{0.32\textwidth}
\centering
\includegraphics[width=\textwidth]{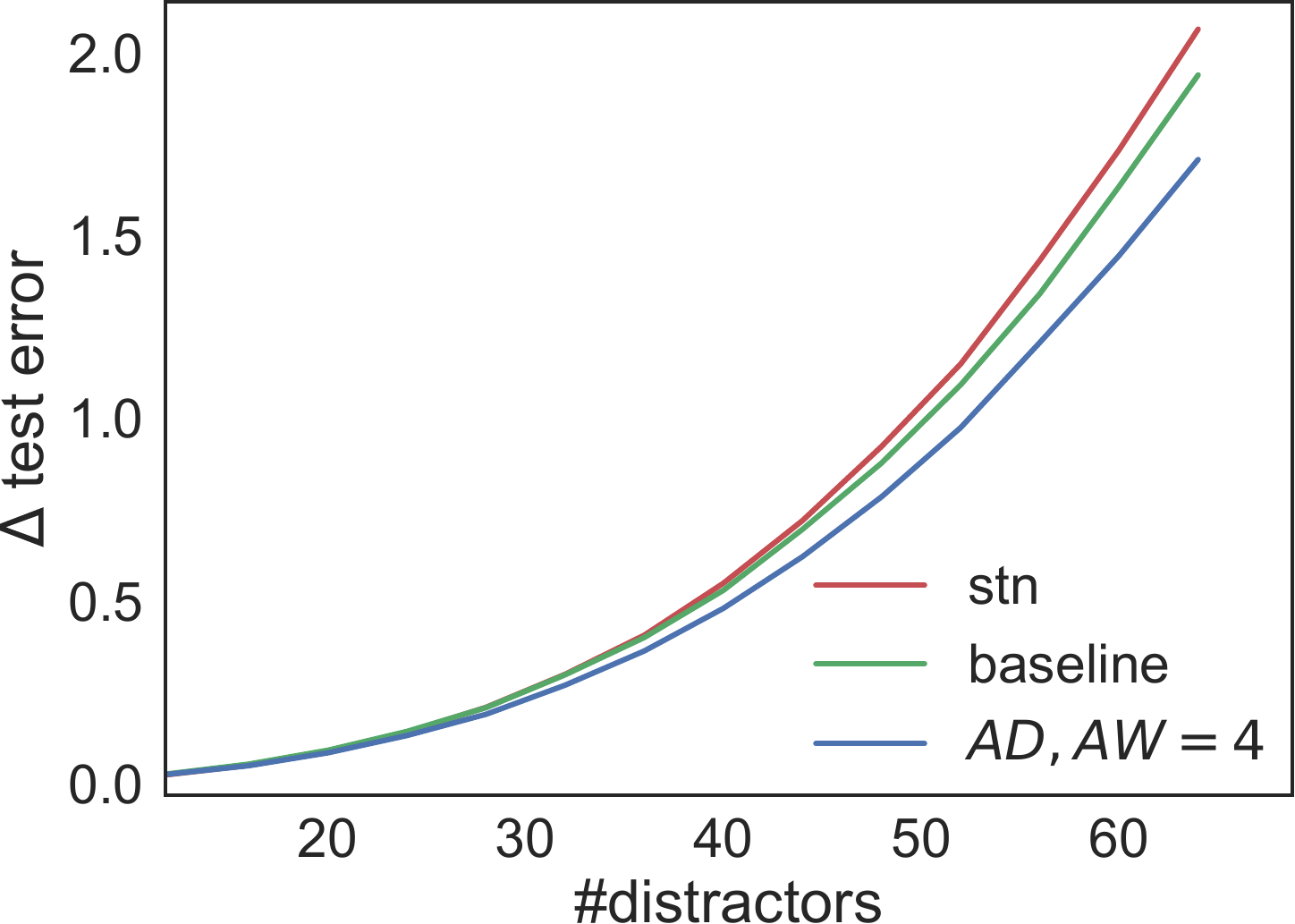}
\caption{Overfitting}
\label{fig:distractors}
\end{subfigure}
\caption{Ablation study on Cluttered Translated MNIST. \texttt{baseline} corresponds to the original model before being augmented with attention. (a) a sample of the cluttered MNIST dataset. (b) the impact of increasing the attention depth (AD), with attention width $AW=1$. (c) effect of increasing AW, for AD=4. (d) best performing model ($AD,AW=4$, softmax attention gates, and regularization \cite{zhao2017diversified}) vs unregularized, sigmoid attention, and without gates. (e) test error of the baseline, attention ($AD,AW=4$), and STN, when trained with different amounts of distractors. }.
\label{fig:ablation}
\end{figure*}

\subsection{Self-attention gates}
Each attention head is enforced to focus on a different region of the image. Thus, it is possible that some of the attention heads attend to regions that are more informative than the ones attended by other heads. To keep the most informative parts and to discard the least informative ones, self-attention \cite{vaswani2017attention} is applied to the different outputs of each attention module $\mathbf{O}^l$:

\begin{equation}
\label{eq:gates}
    \mathcal{S} : \mathbb{R}^{b\times AW \times \#classes} \rightarrow \mathbb{R}^{b \times AW}.
\end{equation}

Similar to $\mathcal{H},\mathcal{O}$, gates are modeled with a $1 \times 1$ convolution kernel $\mathbf{W}_{\mathcal{S}} \in \mathbb{R}^{AW \times c\times 1 \times 1}$:

\begin{equation}
\mathbf{s}^l = softmax(tanh(\sum_{i=0}^{h} \sum_{j=0}^w [ (\mathbf{W_\mathcal{S}}^l \ast \mathbf{Z}^l ) \odot \mathbf{H}_l]_{i,j})),
\end{equation}

where $\mathbf{s}^l$ is a vector containing the gate value of the attention heads at depth $l$ and $\mathbf{W}_\mathcal{S} \in \mathbb{R}^{AW \times c\times 1\times 1}$ is a convolution kernel. We found that using the hyperbolic tangent activation improved the stability at training time since it puts a bound on the softmax input.

The output of each attention module $\hat{\mathbf{o}}^l$ is the weighted average of the prediction of each of its heads:

\begin{equation}
    \hat{\mathbf{o}}^l = \sum_{k=0}^{k=AW} \mathbf{s}_k^l * \mathbf{o}_k^l.
\end{equation}

\subsection{Global attention gates}
The final output of the network consists of the gated average of the outputs of the individual attention modules and the original output of the network. The values of the gates are generated from the last feature map after average pooling $\mathbf{z}^L$, where $L$ is the network depth. Therefore, the model is able to choose the most appropriate regions and levels of abstraction (depth) in order to produce a correct class prediction:

\begin{equation}
    \mathcal{G}: \mathbb{R}^{b\times c} \rightarrow \mathbb{R}^{b \times (AD + 1)},
\end{equation}

where $AD + 1$ corresponds to the dimensionality after concatenating the outputs of each attention module with the original output of the network.

We model this operation with a single linear layer:

\begin{equation}
\label{eq:scores}
\mathbf{g} = softmax(tanh(\mathbf{W_\mathcal{G}}\mathbf{z}^L)),
\end{equation}

with $\mathbf{W_\mathcal{G}} \in \mathbb{R}^{(AD+1) \times c}$ being the weights of the linear operation.

The final prediction is then:

\begin{equation}
\label{eq:all}
\mathbf{output} = g_{orig} \cdot \mathbf{output}_{orig} + \sum_{l\in depths(\mathcal{O})} g^l \cdot \mathbf{\hat{o}}^l,
\end{equation}

where $g_{orig}$ is the gate value for the original network prediction ($\mathbf{output}_{orig}$), $depths(\mathcal{O})$ is the set of depths at which attention is placed, and $\mathbf{output}$ is the final classification corrected by the attentional predictions $\mathbf{\hat{o}}^l$. Note that the prediction of the original pretrained model can be recovered when $\{\mathcal{G} \setminus g_{orig}\} = 0, g_{orig} = 1$.

\section{Experiments}
\label{sec:experiments}
We experimentally prove the repercussion, both in accuracy and consistency, of the different attention modules in our model on the Cluttered Translated MNIST dataset, comparing it with state-of-the-art architectures such as DenseNet \cite{huang2017densely} and ResNeXt \cite{xie2017aggregated}. Finally, we experiment on five fine-grained recognition datasets to demonstrate the universality of our approach.

\subsection{Ablation study}
\label{sect:ablation}
We measure the impact of the different submodules in our approach on the Cluttered Translated MNIST dataset\footnote{\url{https://github.com/deepmind/mnist-cluttered}}. This dataset consists of $40\times 40$ images, each of which has an MNIST \cite{lecun1998mnist} digit in a random position and a set of $D$ randomly placed distractors. The distractors are made with random patches of size $8 \times 8$ from other MNIST digits; a sample is shown in Figure \ref{fig:clutter_mnist}. Following the procedure specified in \cite{mnih2014recurrent}. 
The proposed attention mechanism is used to augment a CNN with five $3\times 3$ convolutional layers and two fully connected layers in the end. The first three convolution layers are followed by batch normalization and a spatial pooling layer. 

The shallowest layers of a CNN, \emph{i.e.}, those closer to the input image, usually perform edge detection and contain very local information about the images. Therefore, attention modules are placed from the deepest levels to the shallowest to obtain higher-level information from local regions of the image. For instance, if the network contains 10 convolution layers and $AD$ is 2, attention is placed on layers 9 and 10, the ones closer to the output. The model is trained for $200$ epochs using SGD with a learning rate of $0.1$, which is annealed by a factor of $0.1$ after epoch $60$. We use a training set of $200k$ images and validation and test sets of $100k$ images each. 
We initialize the weights using the approach from He \emph{et al.} \cite{he2015delving}. The effects of the various tested configurations can be seen in Figure \ref{fig:ablation}, in which the label \texttt{baseline} refers to the performance of the model without attention. Softmax attention gates and regularization \cite{zhao2017diversified} are used in all modules unless explicitly specified.

As shown in Figure \ref{fig:ablation_depth}, $AD$ significantly impacts the accuracy of the model. By adding attention layers with $AW=1$ after each layer, we obtain increasingly better results. When we reach $AD = 4$, the receptive field of the attention module is $5\times5\ px$. Due to the small size of these regions, the performance improvement is saturated at this attention depth.

Once we have determined the best $AD$ and $AW$, \emph{i.e.}, $AD,AW=4$, we perform different experiments with this configuration to test how the softmax compares to sigmoid when applied to the attention masks (Eq. \ref{eq:atthead1}) and how the gates (Eq. \ref{eq:gates}) and the regularization \cite{zhao2017diversified} impact the accuracy of the model. As seen in Figure \ref{fig:hyperparams}, listed by importance, gates, softmax, and regularization are key to increasing the model performance, reaching $97.8\%$. We noticed that gates play a fundamental role in ignoring the distractors; this is especially true for high $AD$ and $AW$.

Lastly, we run our best model so far (Figure \ref{fig:hyperparams}) on the test set, increasing the number of distractors from 4 to 64, to thus ensure that the attention masks are not overfitting the data instead of learning to generalize to any amount of clutter. Additionally, we train the baseline model augmented with an STN \cite{jaderberg2015spatial} and compare their performance cluttered MNIST. Surprisingly, even though the STN model reached comparable performance with our approach in the validation set, the attention augmented model outperforms the other two with a high number of distractors, demonstrating better generalization, as seen in Figure \ref{fig:distractors}.

\subsection{Experiments on CIFAR}
\begin{table}[t!]
\centering
\caption{Number of parameters, floating point operations (Flop), time (s) per epoch on the validation set with batch size 256 on a single GPU, and error rates (\%) on CIFAR. }
\label{tab:performance_benchmark}
\footnotesize
\tabcolsep=0.11cm
\begin{tabular}{@{}lcccccc@{}}
\toprule
 & \textbf{Depth} & \textbf{Params} & \textbf{GFlop} & \textbf{Time} & \textbf{CIFAR-10} & \textbf{CIFAR-100} \\ \midrule
ResNeXt & 29 & 68M & 10.7 & 5.02s & 3.58 & 17.31 \\
Densenet & 190 & \textbf{26M} & 9.3 & 6.41s & 3.46 & \textbf{17.18}\\
WRN & 40 & 56M & 8.1 & 0.18s & 3.80 & 18.30\\ 
WRN-att2 & 40 & 64M & 8.6 & 0.24s & 3.90 & 19.20\\ \midrule
WARN & 28 & 37M & \textbf{5.3} & \textbf{0.17s} & \textbf{3.44} & 18.26\\ 
WARN & 40 & 56M & 8.2 & 0.18s & 3.46 & 17.82\\\bottomrule
\end{tabular}
\end{table}
\begin{figure}[!t]
\centering
\begin{subfigure}[!t]{0.4\textwidth}
\centering
\includegraphics[width=0.9\textwidth]{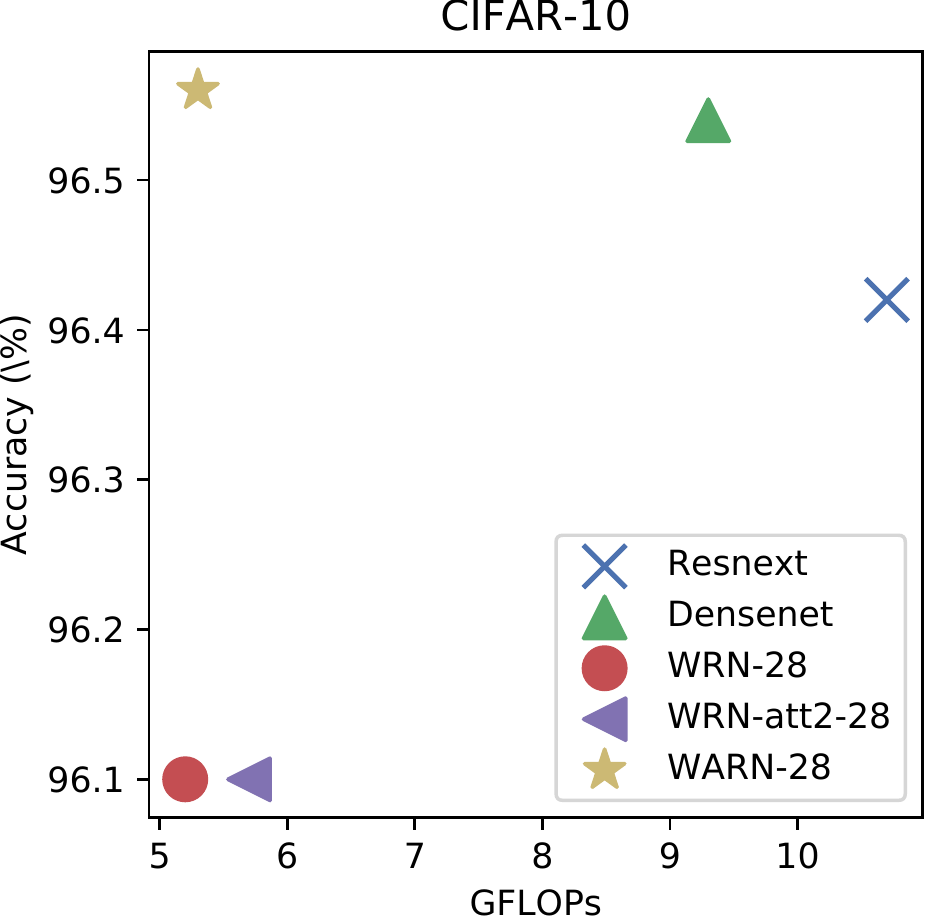}
\caption{CIFAR-10}
\label{fig:cifar_10}
\end{subfigure}
\begin{subfigure}[!t]{0.4\textwidth}
\centering
\includegraphics[width=0.9\textwidth]{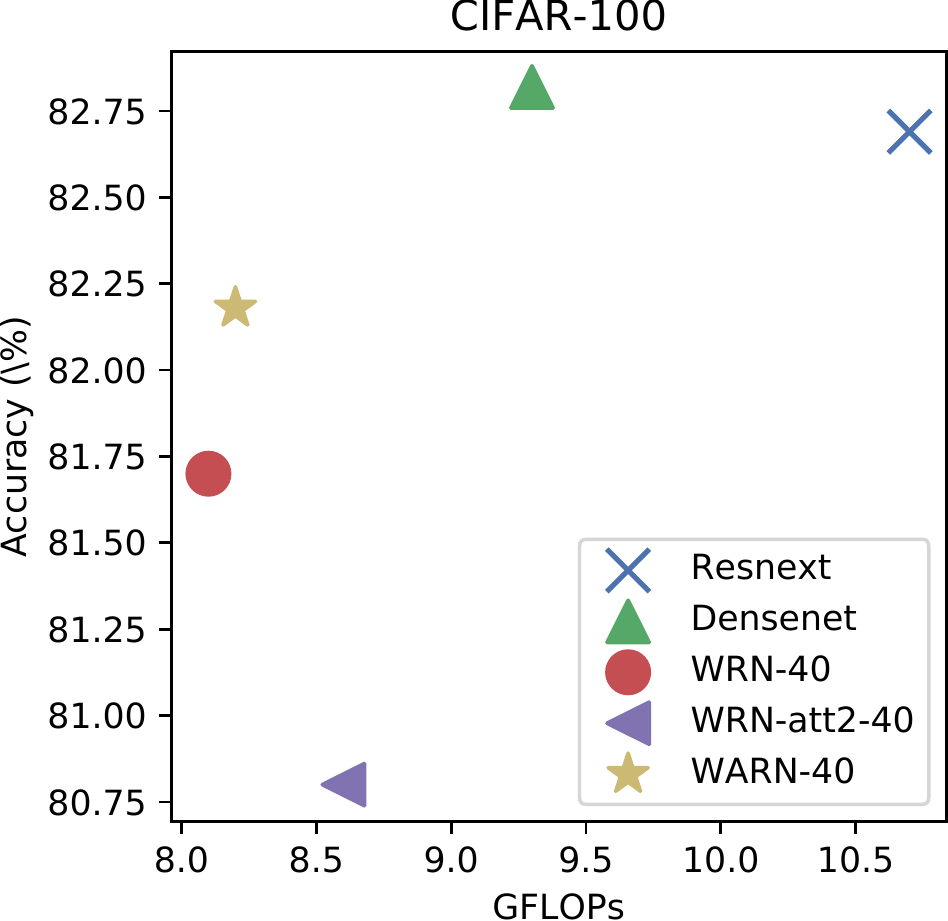}
\caption{CIFAR-100}
\label{fig:cifar_100}
\end{subfigure}
\caption{The performance of ResNeXt, Densenet, WRN, WRN-att2, and WARN on CIFAR-10 and CIFAR-100}
\label{fig:efficiency}
\end{figure}

\begin{table}[t!]
\centering
\caption{Error rate on CIFAR-10 and CIFAR-100 (\%). Results that outperform all other approaches are written in blue, and results that surpass the baseline accuracy are in a black bold font. For fair comparison, the total network depth, attention depth, attention width, the usage of dropout, and the number of floating point operations (Flops) are provided in columns 1-5 }
\label{tab:accuracy-benchmark}
\footnotesize
\begin{tabular}{@{}lccccccc@{}}
\toprule
 & \textbf{Depth} & \multicolumn{1}{c}{\textbf{AD}} & \textbf{AW} & \textbf{Drop} & \textbf{GFlop} & \textbf{C10} & \textbf{C100} \\ \midrule
ResNeXt \cite{xie2017aggregated} & 29 & - & - &  & 10.7 & 3.58 & 17.31 \\ \midrule
\multirow{2}{*}{Densenet \cite{huang2017densely}} & 250 & - & - &  & 5.4 & 3.62 & 17.60 \\
 & 190 & - & - &  & 9.3 & 3.46 & 17.18 \\ \midrule
\multirow{3}{*}{WRN \cite{Zagoruyko2016WRN}} & 28 & - & - &  & 5.2 & 4 & 19.25 \\
 & 28 & - & - & \checkmark & 5.2 & 3.89 & 18.85 \\
 & 40 & - & - & \checkmark & 8.1 & 3.8 & 18.3 \\ \midrule
\multirow{3}{*}{WRN-att2 \cite{jetley2018learn}}  & 28 & 2 & - &  & 5.7 & 4.10 & 21.20 \\ 
& 28 & 2 & - & \checkmark & 5.7 & \textbf{3.60} & 20.00 \\
 & 40 & 2 & - & \checkmark & 8.6 & 3.90 & 19.20 \\ \midrule
\multirow{4}{*}{WARN} & 28 & 2 & 4 &  & 5.2 & \textbf{3.60} & \textbf{18.72} \\
 & 28 & 3 & 4 &  & 5.3 & \textcolor{blue}{3.45} & \textbf{18.61} \\
 & 28 & 3 & 4 & \checkmark & 5.3 & \textcolor{blue}{3.44} & \textbf{18.26} \\
 & 40 & 3 & 4 & \checkmark & 8.2 & \textbf{3.46} & \textbf{17.82}  \\ \bottomrule
\end{tabular}
\end{table}

We compare the proposed attention model with the state of the art on the CIFAR dataset. This dataset contains 50K training images of 32$\times$32 pixels and 10K test images. Images are organized in 10 different categories in the case of CIFAR-10 or 100 in the case of CIFAR-100. Experiments are run with PyTorch \cite{paszke2017pytorch} on 2 NVIDIA 1080Ti and trained following the same procedure as in \cite{Zagoruyko2016WRN}.\footnote{\url{https://github.com/prlz77/attend-and-rectify}}

We chose to augment WRNs \cite{Zagoruyko2016WRN} with the proposed attention mechanism since the number of parameters introduced by our approach is negligible compared to the original number of parameters present in the WRN. We name the resulting model "wide attention residual network" (WARN). Given that the proposed attention mechanism extracts local information at different levels of abstraction and that residual networks only achieve this after convolution groups \cite{veit2018convolutional}, we place the attention modules after each convolution group, from the output of the network to the input image. This approach provides a systematic procedure to place attention on new residual models.

The results are shown in Table \ref{tab:accuracy-benchmark}. We included a WRN augmented with the \texttt{att2} model proposed in \cite{jetley2018learn} for fairness of comparison; we refer to it as WRN-att2. As shown, WARN systematically attains higher accuracy than the baseline architecture (WRN) on CIFAR-10 and CIFAR-100 even without dropout. Interestingly, dropout was required for \cite{jetley2018learn} to obtain a comparable performance, since their approach introduces a larger extra number of parameters. When compared to other architectures, such as Densenet and ResNeXt, WARN achieves state-of-the-art accuracy on CIFAR-10 and is competitive on CIFAR-100 while being up to 36 times faster (see Table \ref{tab:performance_benchmark}). More details on the computing efficiency are shown in Figure \ref{fig:efficiency}. The highest accuracy per Gflop on CIFAR-10 is attained by WARN, and on CIFAR-100, it is more competitive than WRN and WRN-att2.

\begin{table}[!t]
\centering
\caption{Ablation on CIFAR100. The first three columns compare the original gating (softmax or sigmoid) with a simple average or a linear projection. The fourth column shows results w/ and w/o regularization \cite{zhao2017diversified}. The fifth column indicates whether the convnet is updated with the gradients from the attention modules }
\label{tab:cifar100_ablation}
\begin{tabular}{cccccc}
\hline
\multicolumn{3}{c}{\textbf{Gates}} &  &  &  \\
\textbf{Orig} & \textbf{Avg} & \textbf{Linear} & \textbf{Reg} & \textbf{Backprop} & \textbf{Error} \\ \hline
$\checkmark$ &  &  & $\checkmark$ & $\checkmark$ & 18.3 \\
$\checkmark$ &  &  & $\checkmark$ &  & 18.8 \\
$\checkmark$ &  &  &  & $\checkmark$ & 19.2 \\
 & $\checkmark$ &  & $\checkmark$ & $\checkmark$ & 21.5 \\
 &  & $\checkmark$ & $\checkmark$ & $\checkmark$ & 22.5 \\ \hline
\end{tabular}
\end{table}

Finally, we demonstrate that the ablation experiments presented in Section \ref{sect:ablation} are reproducible on CIFAR100. Remarkably, the same conclusions were reached on this dataset. Namely, as we show in Table \ref{tab:accuracy-benchmark}, increasing attention depth and width improved accuracy up to 0.6\%, which was achieved when adding attention to all the residual blocks and increasing the number of attention heads up to four per module. In Table \ref{tab:cifar100_ablation}, we test the effectiveness of the attention gates and regularization. Concretely, we substituted the gates with a fully connected layer that predicted the final probability distribution from the predictions from all the modules, resulting in a 3.4\% error increase. A simple average resulted in a 2.4\% error increase. Regularization provided a 0.3\% accuracy improvement. Finally, we prevented the propagation of gradients from the attention modules to the main architecture to verify that the superior performance of WARN is not caused by optimization effects introduced by the shortcut connection from the loss function to intermediate layers \cite{lee2015deeply}. Although we observed a 0.2\% accuracy drop on CIFAR-10 and 0.5\% drop on CIFAR100, these results are still better than the baseline. Please note that an error increase should be expected since blocking the attention gradients prevents the main architecture from forming adequate features for the attention modules. It is worth noting that in Section \ref{sect:transfer}, WARN improves baseline accuracies even when fine-tuning with the gradients downscaled by a factor of 10 in the main architecture. In contrast, when we performed the same ablation on \cite{jetley2018learn}, it failed to converge since the final prediction completely depends on the predictions from the attended regions.

\begin{figure*}[t!]
\centering
\includegraphics[width=0.75\linewidth]{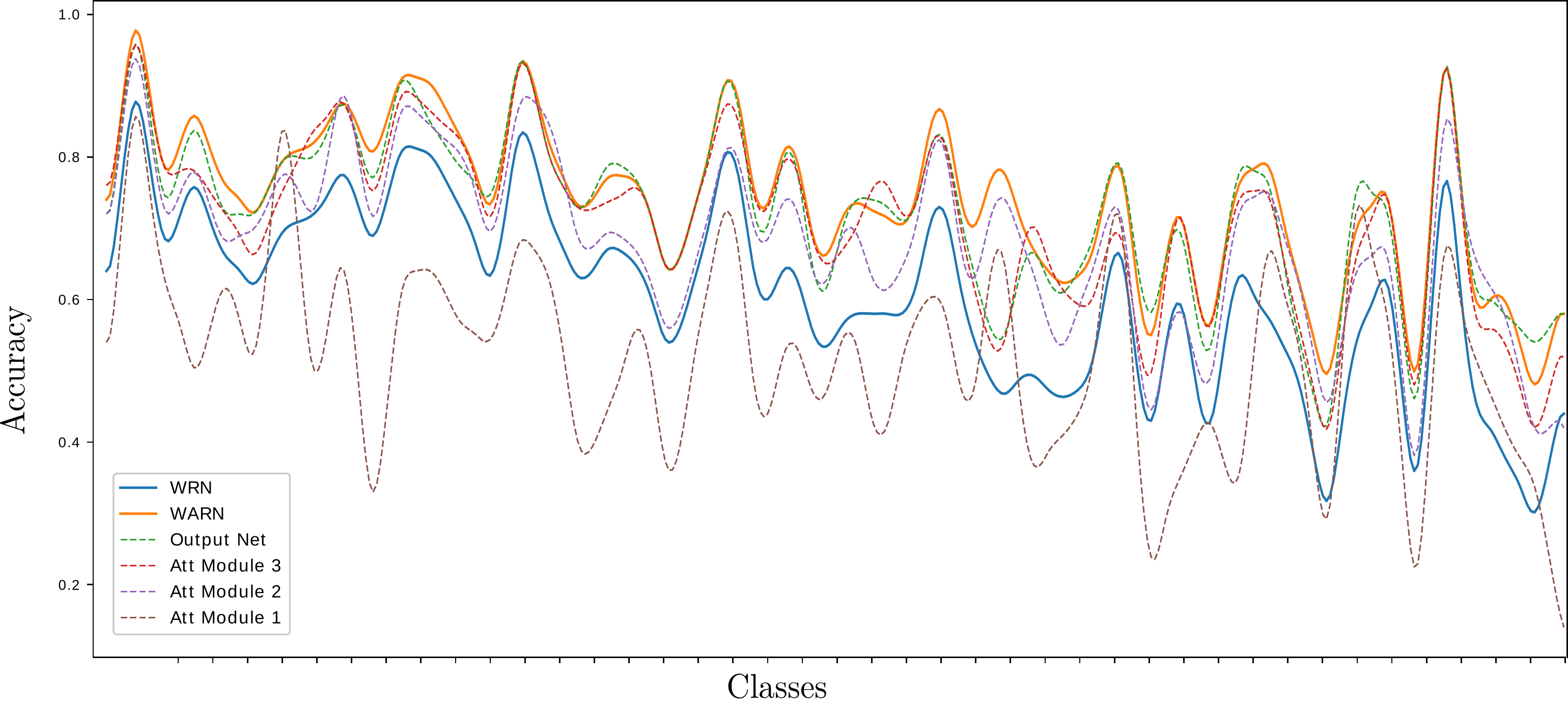}
\caption{ImageNet accuracy for classes most affected by attention. Dashed lines show the accuracy for each of the modules, while solid lines show the final accuracy of the base model and the attention model. Please note that the global gates learn to downscale incorrect predictions, keeping the best overall prediction in most classes.}
\label{fig:module_outputs}
\end{figure*}

\subsection{ImageNet experiments}
\begin{table}[t]
\centering
\caption{ImageNet \cite{russakovsky2012imagenet} validation accuracy (\%) on 224$\times$224 image center crops. The depth of the networks, top-1, and top-5 accuracies are provided in columns 2-3. Bottom rows show networks with our approach. In bold font: the proposed models obtain higher accuracy than the baselines. *PyTorch reimplementation.}
\label{tab:imagenet_results}
\begin{tabular}{@{}lccc@{}}
\toprule

 & \textbf{Top-1} & \textbf{Top-5} \\ \midrule
Attention-56 \cite{wang2017residual} & 78.24 & 94.10 \\
WRN~\cite{Zagoruyko2016WRN} & 78.10 & 93.97 \\
WARN & \textbf{78.29} & \textbf{94.22} \\ 

\midrule
AttentionNext-56 \cite{wang2017residual} & 78.80 & 94.40 \\
ResNeXt-101-64x4* \cite{xie2017aggregated} & 78.80 & 94.40 \\
AResNeXt-101-64x4 & \textbf{79.18} & \textbf{94.55} \\ \bottomrule
\end{tabular}
\end{table}

We verify the scalability of our approach to large-scale multimedia systems on the ImageNet database \cite{russakovsky2012imagenet}, which consists of 1.2M $224\times 224$ images belonging to 1K categories. We use the  training procedure in \href{https://github.com/facebook/fb.resnet.torch}{fb.resnet.torch}. As seen in Table \ref{tab:imagenet_results}, attention consistently improves the performance of wide residual networks \cite{Zagoruyko2016WRN} and ResNeXt \cite{xie2017aggregated} with $AD=3, AW=2$. Residual attention networks \cite{wang2017residual}, a bottom-up approach orthogonal to ours, have also been included for reference. Please note that both approaches could be used at the same time.

\begin{figure*}[!t]
\centering
\includegraphics[width=0.75\textwidth]{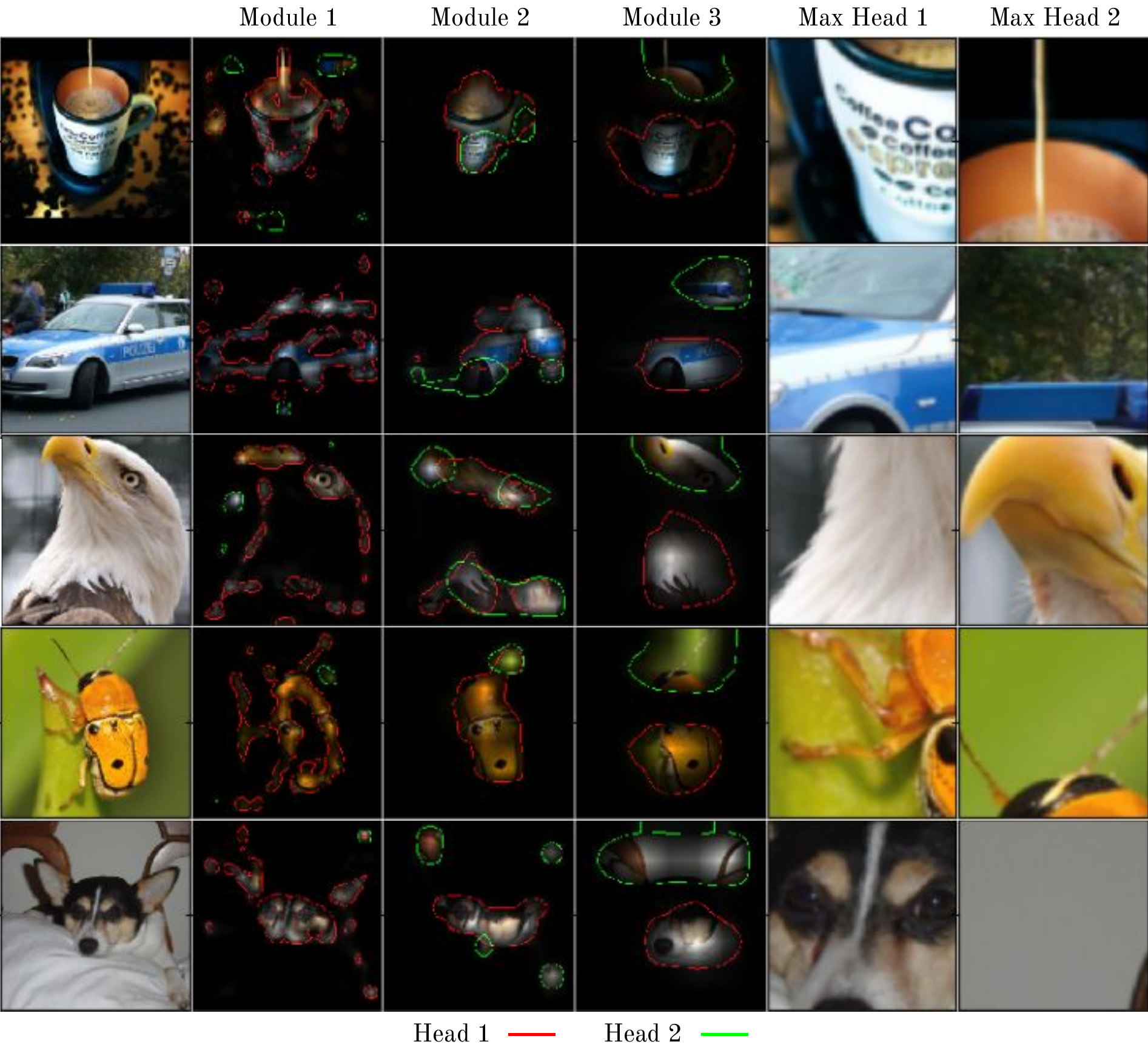}
\caption{Attention masks for five ImageNet classes. Columns 2-4 show the attention masks for each attention module in increasing network depth. Red and green represent different attention heads. The last two columns show patches that maximally activate the attention heads. As shown, attention heads focus on distinct discriminative parts of the foreground object.}
\label{fig:attention_masks}
\end{figure*}

\begin{figure}[!t]
\centering
\includegraphics[width=0.85\linewidth]{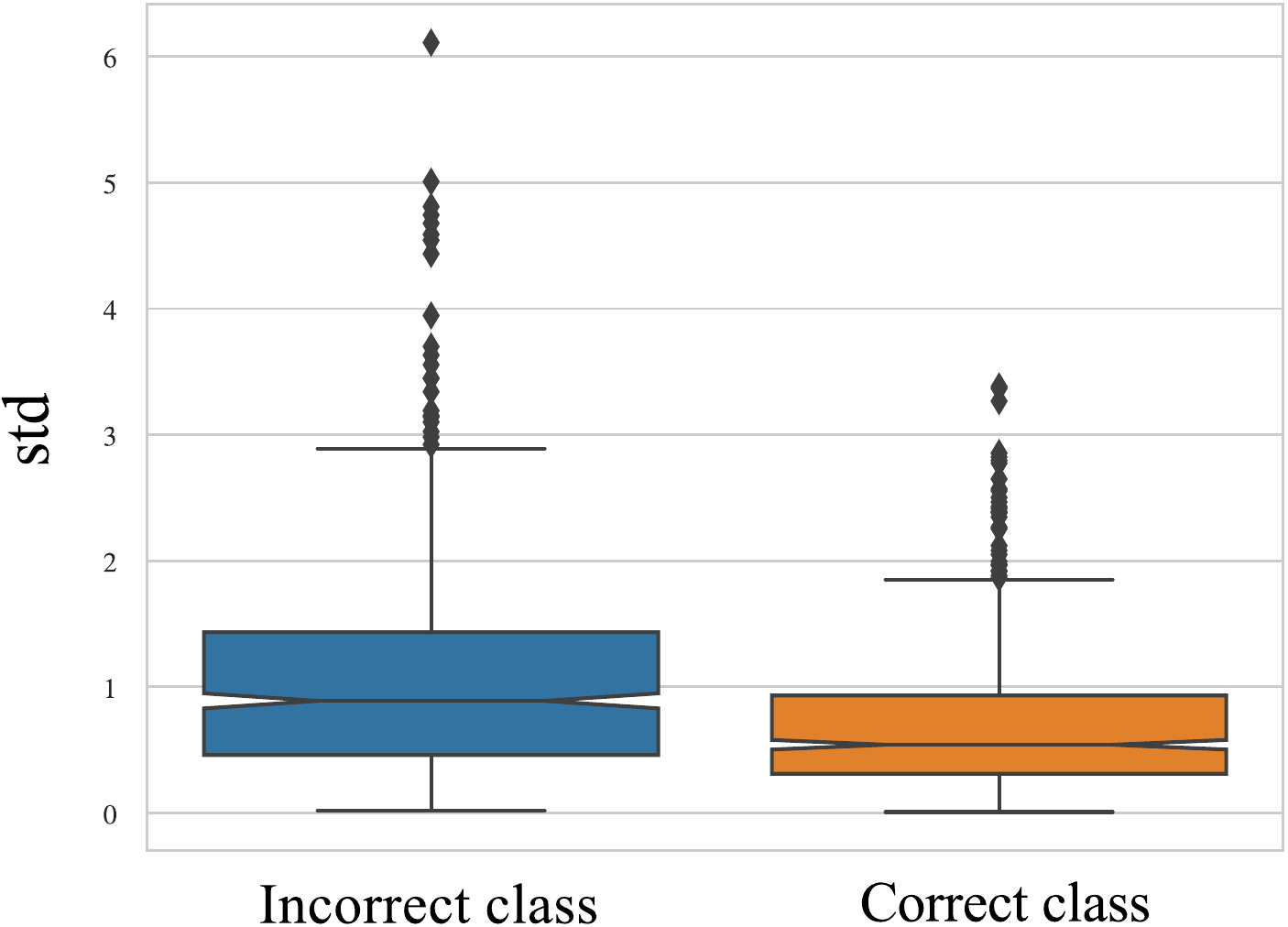}
\caption{Standard deviation per sample of the prediction of the different attention modules for the correct class (orange) and the second most probable prediction (incorrect). Standard deviations are higher for the incorrect classes, indicating that modules tend to disagree when classifying samples incorrectly}
\label{fig:box}
\end{figure}

To provide insight on how attention modules improve the performance of the base architectures, we stored the module predictions of WARN trained on Imagenet and show them in Figure \ref{fig:module_outputs}. As shown, gates successfully choose the best attention module for each class, thus maximizing the final accuracy and obtaining higher scores than the base model without attention (WRN). Interestingly, some of the final accuracies are higher than the accuracy of each individual module. As we show in Figure \ref{fig:box}, this phenomenon is caused by different mistakes in different modules being averaged and thus canceled, thereby acting as an ensemble.

A sample of the attention masks from the network trained on the ImageNet dataset is shown in Figure \ref{fig:attention_masks}. Images were generated using a Canny edge detector over the bilinear interpolation of the attention masks. As shown, attention heads learn to ignore the background and to attend the most discriminative parts of the objects. This result matches the conclusions of section \ref{sect:ablation}. 
For instance, in the second row, it can be seen how the model attends to the top of the police car, focusing on the sirens. In the third row, the model attends to the eagle's neck, probably looking for the distinctive white hair of bald eagles, as well as the shape of the beak. It can also be seen that attention masks do not overlap due to the regularization loss.

\begin{table}[!t]
\centering
\caption{Adaptive computation example. Each row contains results after forcing the network to stop at different depths (first column). As shown, a speedup of 1.6x can be achieved by stopping at the second attention module with a 2.3\% top-5 error increase. Moreover, it is possible to reduce the error increase to 1.6\% when using gate confidences for choosing where to stop computation (adaptive)}
\label{tab:adaptive_computation}
\begin{tabular}{@{}cccc@{}}
\toprule
\textbf{Stop at module} & \textbf{top-1} &   \textbf{top-5} &  
\textbf{Images/s} \\ \midrule
1   & 60.7  & 82.7  & 183.4   \\
2   & 74.9  & 91.9  & 101.8   \\
3   & 77.8  & 93.9  & 63.2    \\
Adaptive & 75.9 & 92.6 & 68.8 \\
None & 78.2 & 94.2  & 62.6    \\ \bottomrule
\end{tabular}
\end{table}

\subsection{Adaptive computation}
\begin{figure}[!t]
\centering
\includegraphics[width=\linewidth]{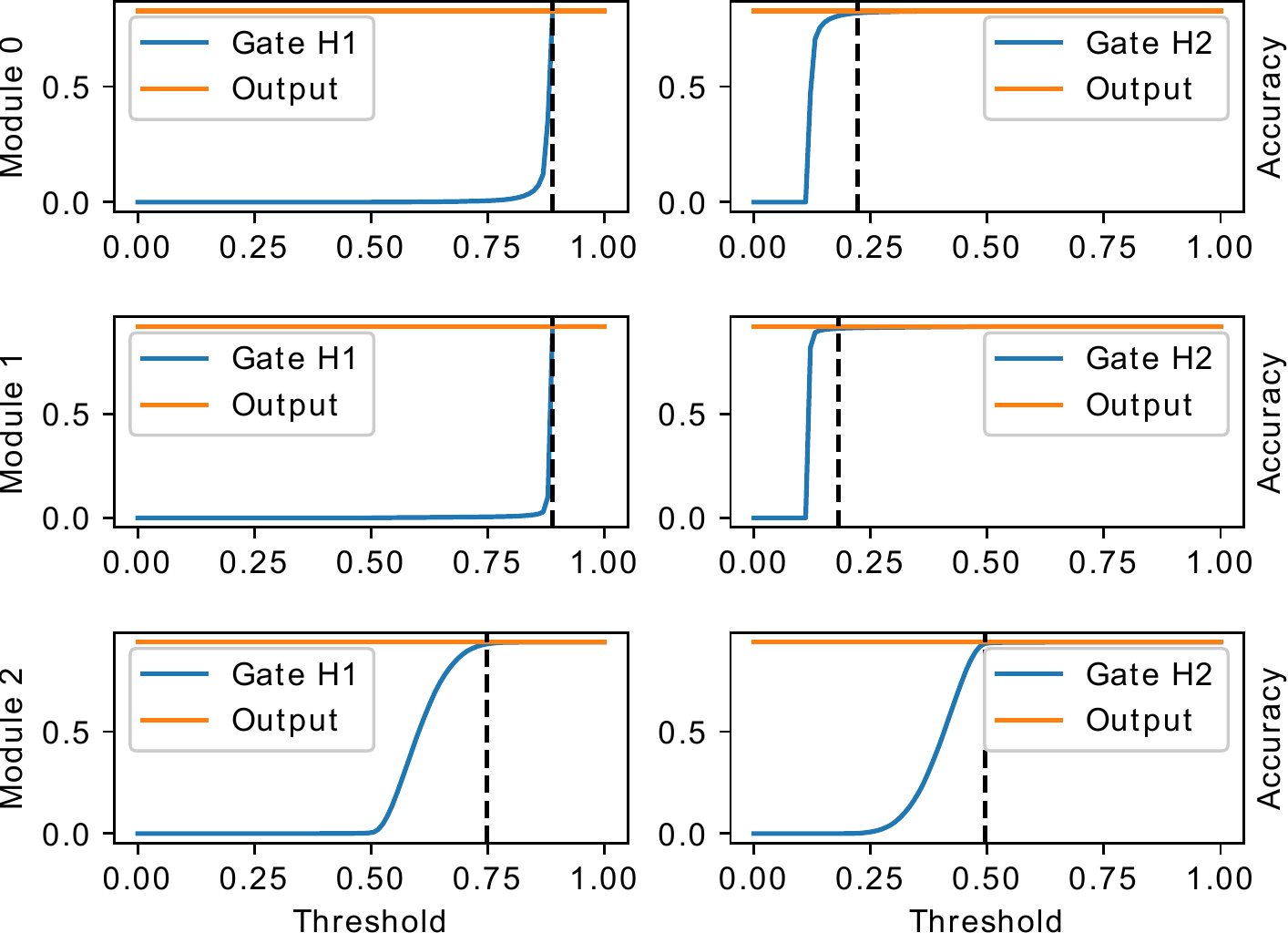}
\caption{Attention head accuracy after using different thresholds on the gate confidence values.}
\label{fig:adaptive_computation}
\end{figure}

An additional advantage of making class predictions at different depths of the network instead of waiting for the output and performing top-down attention is that we can save computation by stopping at a certain attention module before reaching the final output of the network. This is an active research area since it promises more efficient models that need less computational resources and thus are more suited for real-time and mobile applications \cite{zilly2017recurrent,figurnov2017spatially,graves2016adaptive,veit2018convolutional}. Regarding our model, since deeper modules attain higher accuracies, the target is to stop as early as possible, preserving as much accuracy as possible. As a proof of concept, we used the gate values for each of the modules of the WARN trained on Imagenet as a confidence score, thus stopping computation whenever a confidence threshold is reached. Thresholds were chosen according to Figure \ref{fig:adaptive_computation}. The results of this experiment are shown in Table \ref{tab:adaptive_computation}. As shown, without any further training, it is possible to classify images faster with a tradeoff in accuracy. Moreover, the results indicate that this tradeoff could be improved with better stopping policies, for instance, confidence thresholds could be meta-learned \cite{finn2017model}.

\subsection{Transfer Learning}
\label{sect:transfer}
\begin{figure}[t!]
\centering
\begin{subfigure}[t]{0.32\linewidth}
\centering
\includegraphics[width=0.9\textwidth]{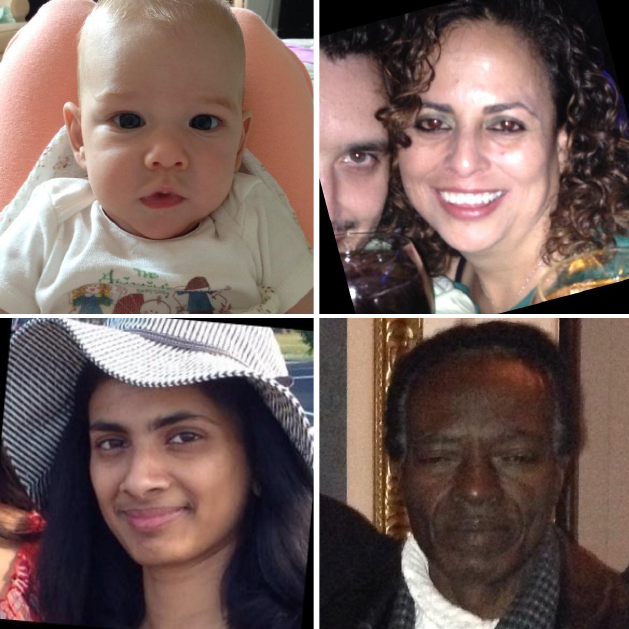}
\caption{}
\label{fig:adience_sample}
\end{subfigure}
\begin{subfigure}[t]{0.32\linewidth}
\centering
\includegraphics[width=0.9\textwidth]{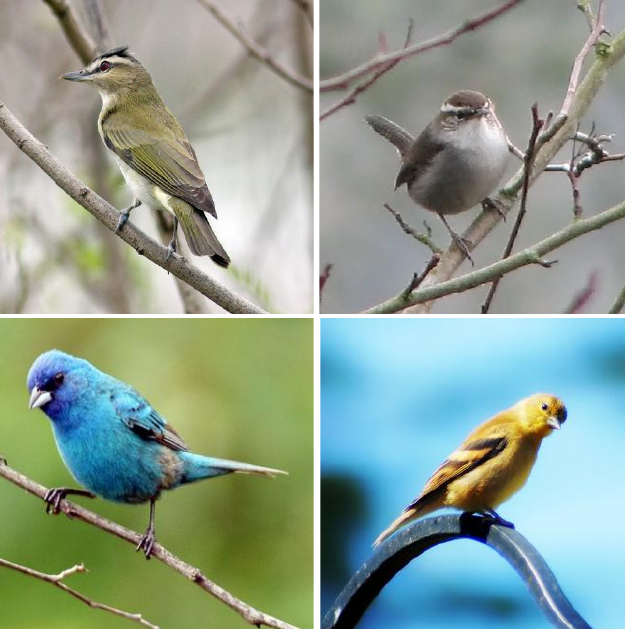}
\caption{}
\label{fig:birds_sample}
\end{subfigure}
\begin{subfigure}[t]{0.32\linewidth}
\centering
\includegraphics[width=0.9\textwidth]{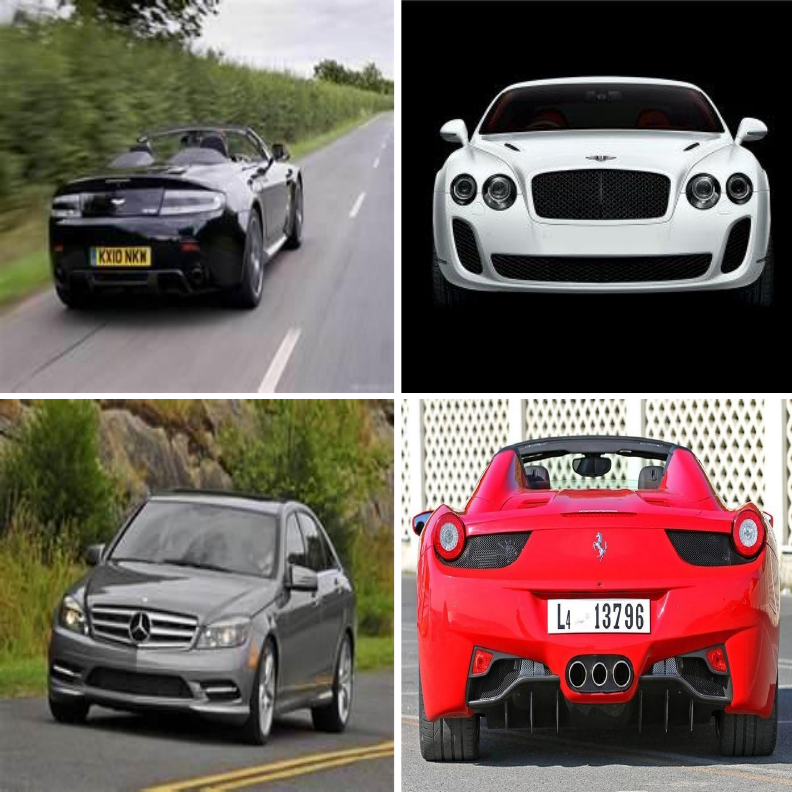}
\caption{}
\label{fig:cars_sample}
\end{subfigure}
\begin{subfigure}[t]{0.32\linewidth}
\centering
\includegraphics[width=0.9\textwidth]{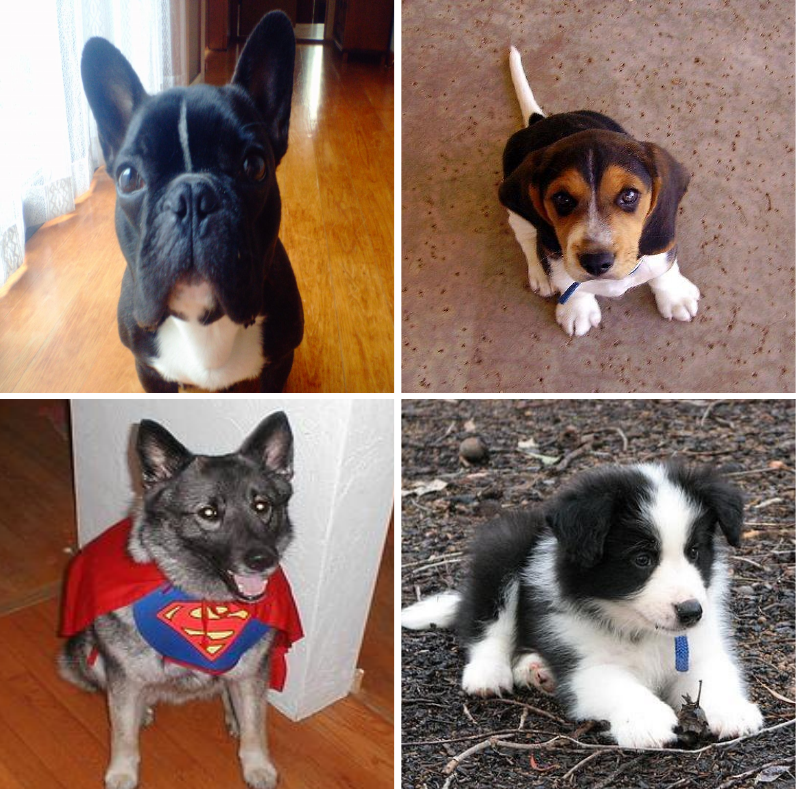}
\caption{}
\label{fig:dogs_sample}
\end{subfigure}
\begin{subfigure}[t]{0.32\linewidth}
\centering
\includegraphics[width=0.9\textwidth]{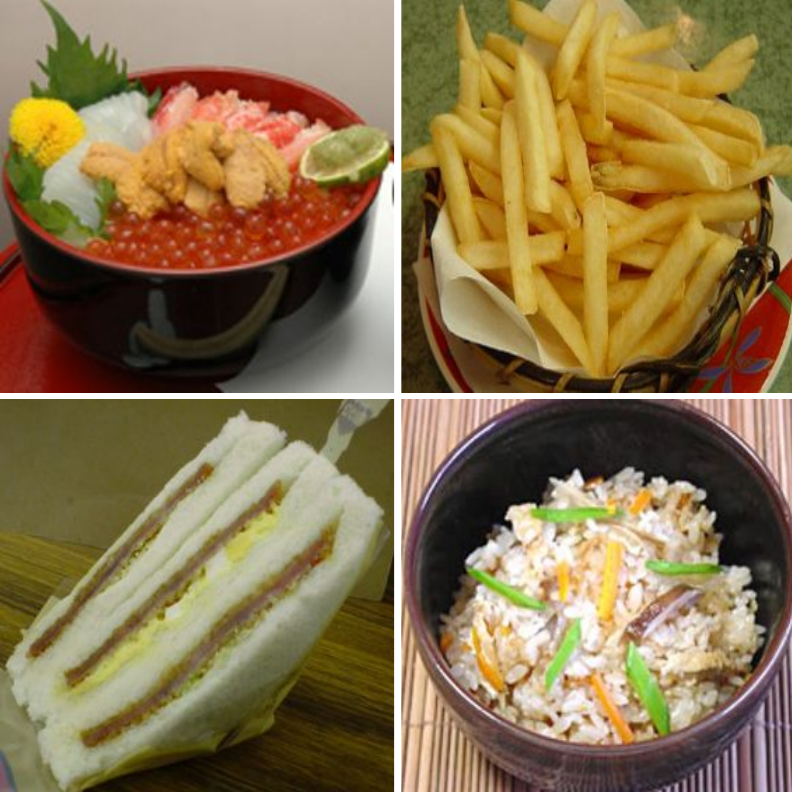}
\caption{}
\label{fig:food_sample}
\end{subfigure}
\caption{Samples from the five fine-grained datasets. (a) Adience, (b) CUB200 Birds, (c) Stanford Cars, (d) Stanford Dogs, (e) UEC-Food100}
\label{fig:dataset-samples}
\end{figure}

We fine-tuned an ImageNet pretrained WRN-50-4 on five fine-grained recognition datasets: Stanford Dogs \cite{khosla2011novel}, UEC Food-100 \cite{matsuda12}, Adience \cite{eidinger2014age}, Stanford Cars \cite{krause20133d}, and CUB200-2001 \cite{WahCUB_200_2011}; see Figure \ref{fig:dataset-samples} for a sample from each dataset. Fine-tuning was performed for 100 epochs with SGD and early stopping. The batch size was set to 64, and the initial learning rate was set to $10^{-3}$ for the main architecture and $10^{-2}$ for the final classifier and the attention layers. The learning rate was decayed by a factor of $10$ at 30, 60, and 90 epochs. Standard data augmentation was used, \emph{i.e.}, $224\times 224$ crops from $256 \times 256$ images and random horizontal flip. Please note that other approaches such as RA-CNNs or MA-CNN make use of images at higher resolution, $512 \times 512$, or color jitter \cite{hassannejad2016food}, which is essential to achieve state-of-the-art performances. However, the aim of this experiment is to show that the proposed mechanism consistently achieves higher accuracy than the baseline model for a wide variety of fine-grained recognition tasks. Notwithstanding, WARN-50-4 achieved state-of-the-art performance for Adience Gender, Stanford Dogs and UEC Food-100.

\begin{table*}[!t]
\centering
\caption{Results on six fine-grained recognition tasks. \emph{DSP} means that the cited model uses domain-specific pretraining. \emph{HR} means the cited model uses high-resolution images. Models that use attention are marked with ($^*$). Accuracies that improve the baseline model are in black bold font, and state-of-the-art accuracies are in blue}
\label{tab:fine-grained-results}
\begin{tabular}{@{}l|cc|cccccc@{}}
\toprule
 & HR & DSP & \textbf{Dogs} & \textbf{Cars} & \textbf{Birds} & \textbf{Food} & \textbf{Age} & \textbf{Gender} \\ \midrule
Sermanet \emph{et al.} \cite{sermanet2014attention} & Y & N & 76.8 & - & - & - & - & - \\
RA-CNN \cite{fu2017look} & Y & N & 87.3 & 92.5 & 85.3 & - & - & - \\
DVAN \cite{zhao2017diversified} & N & N & 81.5 & 87.1 & 79.0 & - & - & - \\
MA-CNN \cite{zheng2017learning} & Y & N & - & 92.8 & 86.5 & - & - & - \\ 
Wu \emph{et al.} \cite{wu2018deep} & Y & N & - & 93.4 & 89.7 & - & - & - \\
NTS-Net \cite{yang2018learning} & Y & N & - & 93.9 & 87.5 & - & - & - \\
DLA \cite{yu2018deep} & Y & N  & - & 94.1 & 85.1 & - & - & - \\
Inception \cite{hassannejad2016food} & N & N & - & - & - & 81.5 & - & - \\
FAM \cite{rodriguez2017age} & N & Y & - & - & - & - & 61.8 & 93.0 \\
DEX \cite{Rothe-IJCV-2016} & N & Y & - & - & - & - & 64.0 & - \\ \midrule
WRN \cite{Zagoruyko2016WRN} & N & N & 89.6 & 88.5 & 84.3 & 84.3 & 57.4 & 93.9 \\
WARN* & N  & N & \textcolor{blue}{92.9} & \textbf{90.0} & \textbf{85.6} & \textcolor{blue}{85.5} & \textbf{59.7} & \textcolor{blue}{94.6} \\ \bottomrule
\end{tabular}
\end{table*}

In particular, in Table \ref{tab:fine-grained-results}, we show that fine-tuning with WARN substantially increases the baseline accuracy on all benchmarks. Moreover, competitive performance was obtained even when comparing our model with approaches that use a higher resolution \cite{yang2018learning} or domain-specific pretraining \cite{Rothe-IJCV-2016}. For instance, in DEX \cite{Rothe-ICCVW-2015}, CNN was pretrained with millions of faces for age and gender recognition. Differently, we were able to report state-of-the-art accuracies on Stanford Dogs, UEC Food, and the Adience Gender recognition benchmark and competitive scores on the other three tasks while using low-resolution images with standard ImageNet pretraining. Please note that the $+1.3\%$ accuracy increase reported on CUB200-2011 is higher than the improvement reported by STNs \cite{jaderberg2015spatial} ($+0.8\%$), even though we augment a stronger baseline, and thus, diminishing returns should be expected. This result indicates that while STN discards information from the main architecture, our model is able to extract additional information that improves the final performance and does it with a high efficiency per introduced parameter; see Table \ref{tab:acc_param}.

\section{Conclusions}
Attention has emerged as an effective mechanism to improve the precision and efficiency of multimedia systems by focusing on the most informative data while discarding clutter and noise. Inspired by the human visual system, two kinds of attention mechanisms have emerged in the literature: top-down iterative processes to choose the relevant regions from global information about the scene and bottom-up approaches, which detect the most relevant and salient regions along the visual path \cite{le2013visual}.

However, due to their iterative nature, top-down processes \cite{chen2017sca,sharma2015action,zheng2017learning, wu2018deep} are slower than single-pass bottom-up approaches and are more difficult to train, especially those using reinforcement learning. In addition, these kinds of models depend on a correct understanding of the "world" (the context being attended) to propose meaningful regions to focus on. Moreover, in humans and animals, there are complex interactions with the observer's motivation and experience \cite{le2013visual}. On the other hand, bottom-up approaches \cite{xiao2015application, peng2018object, jaderberg2015spatial, jetley2018learn} select the most important regions from the input, sequentially or as a previous stage. However, errors produced in these sequential processes amplify with depth (for instance, focusing on a distractor in the first stages of the network), and they also introduce an additional overhead, either with extra layers or detection algorithms.

\begin{table}[!t]
\centering
\caption{Increment of accuracy (\%) per million parameters}
\label{tab:acc_param}
\footnotesize
\tabcolsep=0.11cm
\begin{tabular}{@{}lccccccc@{}}
\toprule
\textbf{} & \textbf{Dogs} & \textbf{Food} & \textbf{Cars} & \textbf{Gender} & \textbf{Age} & \textbf{Birds} & \textbf{Average} \\ \midrule
WRN & 1.3 & 1.2 & 1.3 & 1.4 & 0.8 & 1.2 & 1.2 \\
WARN & \textbf{6.9} & \textbf{2.5} & \textbf{3.1} & \textbf{1.5} & \textbf{4.0} & \textbf{2.5} & \textbf{3.4} \\ \bottomrule
\end{tabular}
\end{table}

Differently, we propose a bottom-up approach that runs in parallel to the visual path, without discarding any relevant information and with negligible computational overhead. Note that after this parallel process, sequential attention can still be applied, similar to the human visual system \cite{le2013visual}.

The proposed model outperforms the accuracy of DenseNet and ResNeXt on CIFAR-10 while being 37 times faster than the former and 30 times faster than the latter. We have demonstrated the scalability of our model with experiments on ImageNet, improving top-5 accuracy by 0.15\% on both ResNeXt and WRN, thus proving the universality of our model to extend different architectures. In transfer learning experiments, we demonstrated that the proposed model can be used to extend pretrained networks, improving the accuracy of WRN on six different fine-grained recognition tasks and achieving state-of-the-art results on Stanford Dogs, UEC Food-100, and the Adience Gender classification benchmark. In analytical and qualitative experiments, we have demonstrated that the proposed mechanism corrects misclassifications from the original architecture by attending local discriminative regions. Finally, we have shown that adaptive computation time is a promising future direction for this work.



%


\section*{Acknowledgments}
The authors acknowledge the support of the Spanish project TIN2015-65464-R (MINECO/FEDER), the 2016FI B 01163 grant of Generalitat de Catalunya, and the COST Action IC1307 iV\&L Net. We also gratefully acknowledge the support of NVIDIA Corporation with the donation of a Tesla K40 GPU and a GTX TITAN GPU used for this research.

{\small
\bibliographystyle{ieee_fullname}
\bibliography{egpaper.bbl}
}

\end{document}